\documentclass[journal]{IEEEtran}
\ifCLASSINFOpdf
\else
\fi

\usepackage{subfigure}
\usepackage{epsfig}
\usepackage{graphicx}
\usepackage{amsmath}
\usepackage{amssymb}
\usepackage{algorithm, algorithmic}
\usepackage{multirow}
\newcommand{\tabincell}[2]{\begin{tabular}{@{}#1@{}}#2\end{tabular}}
\usepackage{url}
\usepackage{color}
\begin{document}
%
\title{Deep Manifold Embedding for Hyperspectral Image Classification}
%
%
%

\author{Zhiqiang~Gong,
        Weidong~Hu,
        Xiaoyong~Du,
        Ping~Zhong,~\IEEEmembership{Senior Member,~IEEE}
        and~Panhe~Hu
\thanks{Manuscript received XX, 2020; revised XX, 2020. This work was supported by the Natural Science Foundation of China under Grant 61671456, 61971428, and 62001502. (Corresponding author: Ping Zhong)}
\thanks{Z. Gong 
is with the National Innovation Institute of Defense Technology,
Chinese Academy of Military Science, Beijing 100000, China. e-mail: gongzhiqiang13@nudt.edu.cn.}
\thanks{ W. Hu, X. Du, P. Zhong, and P. Hu are with the National Key Laboratory of Science and Technology on ATR, College of Electrical Science and Technology, National University of Defense Technology, Changsha, China, 410073. e-mail: (wdhu@nudt.edu.cn, xydu@nudt.edu.cn, zhongping@nudt.edu.cn, hupanhe13@nudt.edu.cn).}
}

%
%

\markboth{IEEE LATEX,~Vol X, 2020}%
{Shell \MakeLowercase{\textit{et al.}}: Bare Demo of IEEEtran.cls for IEEE Journals}
%



\maketitle

\begin{abstract}
 Deep learning methods have played a more and more important role in hyperspectral image classification. However, general deep learning methods mainly take advantage of the sample-wise information to formulate the training loss while ignoring the intrinsic data structure of each class. Due to the high spectral dimension and great redundancy between different spectral channels in hyperspectral image, these former training losses usually cannot work so well for the deep representation of the image.
To tackle this problem, this work develops a novel deep manifold embedding method (DMEM) for deep learning in hyperspectral image classification.
First, each class in the image is modelled as a specific nonlinear manifold and
 the geodesic distance is used to measure the correlation between the samples.
Then, based on the hierarchical clustering, the manifold structure of the data can be captured and each nonlinear data manifold can be divided into several sub-classes.
Finally, considering the distribution of each sub-class and the correlation between different sub-classes under data manifold, the DMEM is constructed as the novel training loss to incorporate the special class-wise information in the training process and obtain discriminative representation for the hyperspectral image.
Experiments over  four real-world hyperspectral image datasets have demonstrated the effectiveness of the proposed method when compared with general samples-based losses and also shown the superiority when compared with the state-of-the-art methods.
\end{abstract}


%
\IEEEpeerreviewmaketitle

\section{Introduction}

Recently, hyperspectral images, which contain hundreds of spectral bands to characterize different materials, make it possible to discriminate different objects with the plentiful spectral information and have proven its important role in the literature of remote sensing and {computer vision \cite{sun2020,feng2019, luo2018}}. As an important hyperspectral data {processing} task, hyperspectral image classification aims to assign the {unique land-cover label to each pixel \cite{sun2019,zhan2019}} and is also the key technique in many real-world applications, such as the urban planning \cite{urban}, military applications \cite{military}, and others.
However,
hyperspectral image classification is still a challenging task.
There exists high nonlinearity of samples within each class due to the high spectral channels, which makes the representation under Euclidean distance cannot work well. Therefore, how to effectively model and represent the samples of each class tends to be a difficult problem.
Besides, great overlapping which occurs between the spectral {bands} from different classes in the hyperspectral image \cite{manolakis2003,gong_statistical}, multiplies the difficulty to obtain discriminative features from the samples.

Deep models have demonstrated their potential to model and represent the samples in various tasks, including the field of hyperspectral image classification \cite{deep_1,deep_2}.
It can learn the model adaptively with the data information from the training samples and extract {discriminative features} between different classes.
Due to the good performance, this work takes advantage of the deep model to extract features from the hyperspectral image.
However, large amounts of training samples are required {for the training of} the deep model while there usually exists limited number of {training samples in the literature of hyperspectral image classification \cite{gong_statistical}}. Therefore, how to construct the training loss and fully utilize the data information with a certain limited number of training samples becomes the essential and key problem for {training deep model effectively}.

General deep learning methods mainly construct the samples-wise losses for the training process.
The softmax loss, namely the softmax cross-entropy loss, is widely applied in prior works. It is formulated by the cross entropy between the posterior probability and the class label of each sample \cite{gaussian_mixture_loss}, which mainly takes advantage of the point-to-point information of each sample itself. Several {variants try to utilize the inter-sample information between sample pairs or among sample triplets.}
These losses, such as the contrastive loss \cite{contrastive_loss} and triplet loss \cite{triplet_loss}, have made great strides in improving the representational ability of the CNN model.
However, these prior losses mainly utilize the data information of sample itself or between samples and ignore the intrinsic {data structure of each class}. In other words, these samples-wise methods only consider the commonly simple information and ignore the special intrinsic data structures within each class {in the training process}.

Due to the high spectral dimensions and {great overlapping between spectral bands} in hyperspectral image, these samples-wise losses in prior works usually cannot be well fit for hyperspectral image classification. Therefore, this work {attempts to} take advantage of the {class-wise information of hyperspectral image based on these spectral characteristics} and construct the specific loss for better deep representation {of the image}.
{The key process is to model intrinsic {data structure of each class in the} hyperspectral image.}
Considering the characteristics of hyperspectral image, nonlinear manifold model, which plays an important role in the nonparametric models, is applied to model the image for the current task.

\begin{figure}[t]
\centering
 \subfigure[]{\label{fig:class_pavia}\includegraphics[width=0.48\linewidth]{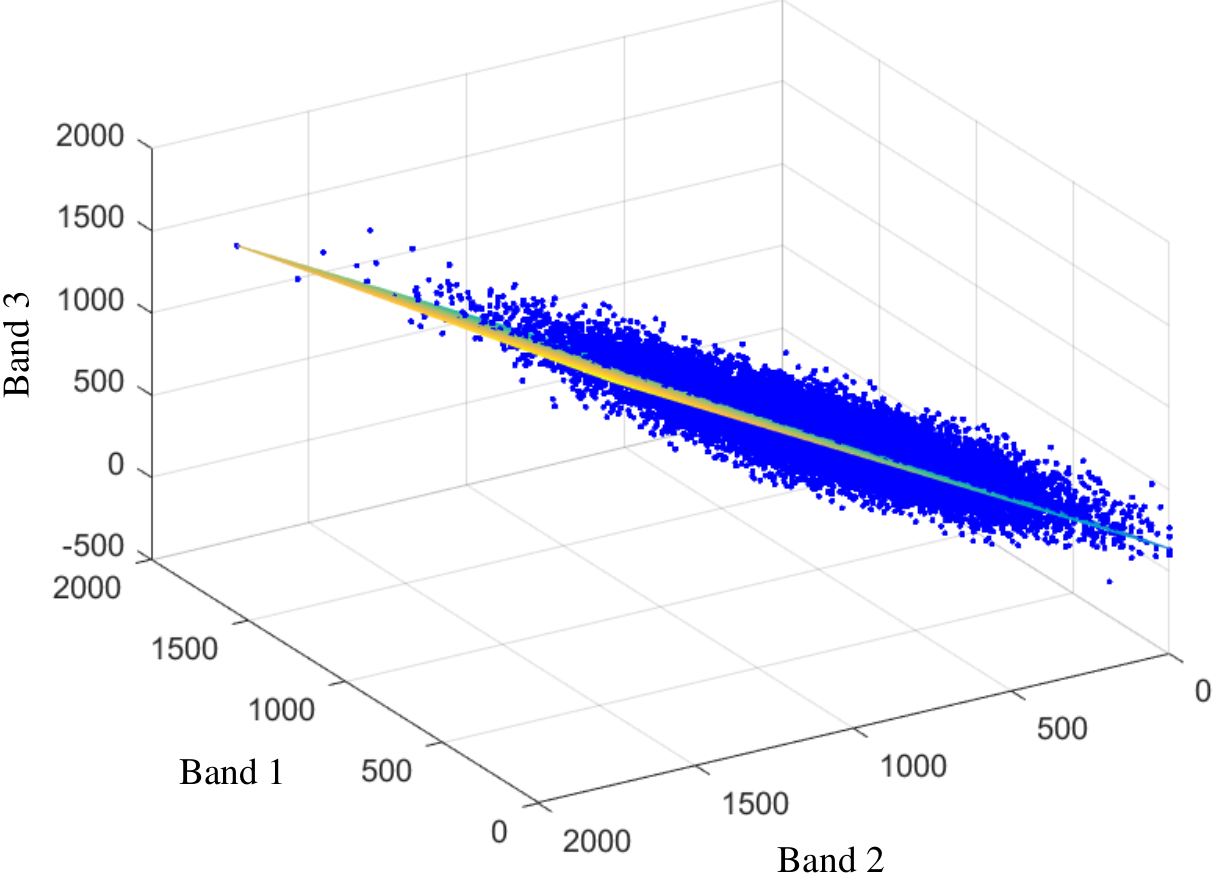}}
 \subfigure[]{\label{fig:class_salinas}\includegraphics[width=0.48\linewidth]{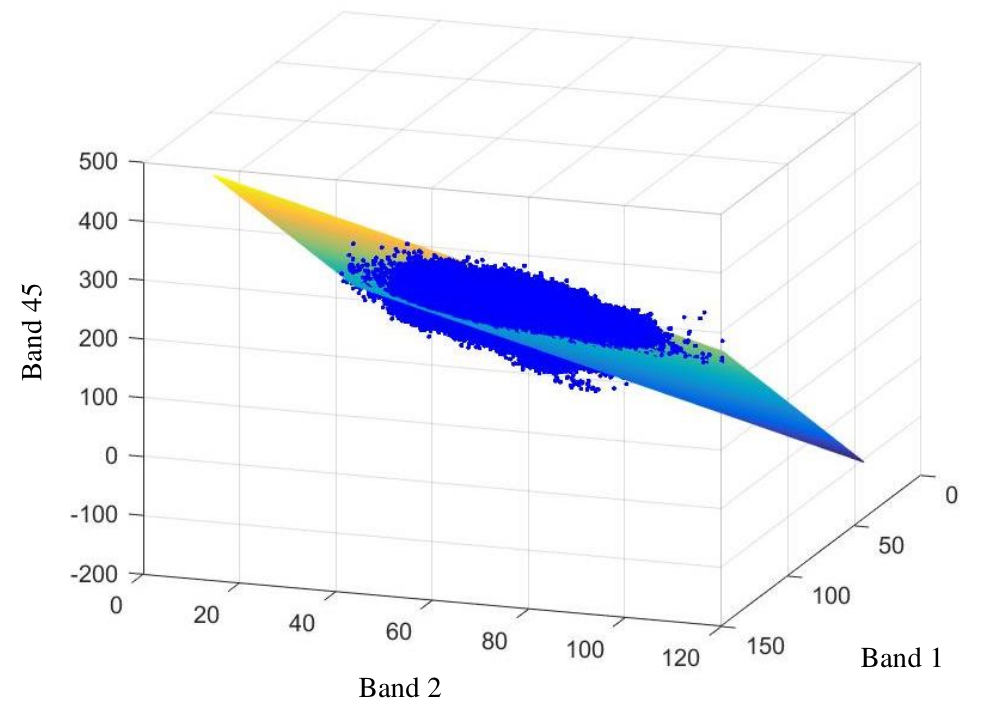}}
   \caption{Manifold structure of different classes in (a) Pavia University; (b) Houston2018 data. In the figure, three bands have been selected and samples from a given class in different datasets have been visualized in 3-D space(e.g., meadows in Pavia University, non-residential buildings in Houston 2018 data).}
\label{fig:manifold_structure}
\end{figure}

Generally, a data manifold follows the law of manifold distribution: in real-world applications, high-dimensional data of the same class usually draws close to a low dimensional manifold \cite{manifold_principle}. Therefore, hyperspectral images, which provide a dense spectral sampling at each pixel, possess good intrinsic manifold structure.
{For better understanding, we select three bands from the hyperspectral image and visualize the samples from a given class in three-dimensional space. As Fig. \ref{fig:manifold_structure} shows, the samples from each dataset are approximately fit to a specific two-dimensional plane. Based on the definition of manifold where a manifold of dimension $n$ is a set of points that is homeomorphic to $n$-dimensional Euclidean space, these samples in hyperspectral image can be modelled as a low-dimensional manifold in high-dimensional space.
Therefore, this} work aims to develop a novel manifold embedding method in deep learning (DMEM) for hyperspectral image classification to make use of the data manifold structure and  preserve the {intrinsic data structure of each class} in the obtained low dimensional features.

In addition to the law of manifold distribution, data manifold also follows another law of cluster distribution: {A certain class of the high-dimensional data can be divided into several subclasses where each subclass correspond to a specific probability distributions on the manifold} \cite{manifold_sub_principle}. {Under the geodesic distances between the samples,  this work divides each class of the hyperspectral image into several sub-classes.} {While in the feature space, these probability distributions are expected to be far enough to distinguish these subclasses.}
Therefore, we develop the DMEM for deep learning following two principles.
\begin{enumerate}
  \item Based on multi-statistical analysis, deep manifold embedding can be constructed to encourage the features from each sub-class to follow a certain statistical distribution and further preserve the intrinsic statistical structure in the low dimensional feature space.
  \item Motivated by the idea of maximizing the ``manifold margin'' {in} manifold discriminant analysis \cite{manifold_distance}, additional diversity-promoting term is developed to increase the margin between sub-classes from different data manifold.
\end{enumerate}

Overall, the main contributions of this work are threefold. Firstly, this work models the hyperspectral image with the nonlinear manifold  and takes advantage of the intrinsic manifold structure of the hyperspectral image in the deep learning process. Secondly, this work formulates a novel training loss based on the manifold embedding in deep learning for hyperspectral image classification and thus {the class-wise information} can be fully utilized in the deep learning process.
To the authors' best knowledge, this paper first develops the specific training loss using the manifold structure of the hyperspectral image.
Finally, a thorough comparison is provided using the existing samples-based embedding and losses.

The rest of the paper is organized as follows. We review the related works {to our approach} in Section \ref{sec:related}. Section \ref{sec:manifold_embedding} details the construction of {the proposed DMEM} for hyperspectral image classification. In Section \ref{sec:experiment}, experiments on {four real-world} hyperspectral images demonstrate the effectiveness of the proposed method. Finally, Section \ref{sec:conclusion} summarizes this paper with some discussions.

\section{Related Work}\label{sec:related}
In this section, we will review two topics that closely related to this paper. First, deep learning methods for hyperspectral image classification are briefly introduced, which promote the generation of motivations of this work.
Then, manifold learning in prior works is investigated, which is directly the related work of the proposed method.

\subsection{Deep Learning for Hyperspectral Image Classification}\label{subsec:deep_learning}
General deep learning methods for hyperspectral image classification mainly introduce existed samples-based losses or their variants for the training process. These loss functions can be divided into two classes according to different criterions.

{Most of the prior works utilize the simple one-to-one correspondence criterion, which measures the difference between the predicted and the corresponding label of each sample, for the training of the deep model  in hyperspectral image classification \cite{paolettime2019}.
These works \cite{paoletti2019, zhongzl2017} mainly focus on the design of network architectures to {capture spectral signature, such as Convolutional Neural Networks (CNNs) \cite{gong_1} and Recurrent Neural Networks (RNNs) \cite{liu2017, hang2019, zhou2019} or the combination} with other methods, such as Conditional Random Field (CRF) \cite{alam2018}, to capture additional spatial information. Therefore, the commonly used softmax loss is usually adopted for the training process.}

Due to the high intraclass variance and low interclass variance within the hyperpectral image, many works take advantage of the inter-sample information for the training of {the} deep model.
Generally, these works use metric learning in the {training process} for hyperspectral image classification. The principle is to decrease the Euclidean distances {between samples} with the same class label while {increase} the distances of samples from different classes \cite{gong_2}.
{For example, Guo \textit{et al.} \cite{guo2019} introduces center loss \cite{center} in hyperspectral image classification which utilizes the center point of each class to formulate the image pairs within the class.
Our prior work \cite{gong_1} uses and further improves the structured loss \cite{lifted} which takes advantage of the intrinsic structure within the mini-batch.}
Liang \textit{et al.} \cite{liang2019} combines general pairwise information with conditional random field. These losses for the training of deep model consider the inter-sample information and take advantage of more useful information in the training process, and therefore the learned model can obtain a better representation for hyperspectral image.

However, the former losses only consider the commonly existed information from the training samples and ignore the intrinsic structure information of each class. Especially, for the task at hand, there exist high nonlinearity and great overlapping of the spectral signature {of each class in the high} dimensional hyperspectral data. These {class information} cannot be fully utilized with {the existed samples-based} training losses.
Under these circumstances, constructing a novel training loss for hyperspectral image which can be better fit for the hyperspectral image would be particularly important in the deep representation of the image. 
This is also the direct motivations of the developed method in this work.

\subsection{Manifold Learning}\label{subsec:manifold_learning}
Manifold learning is the research topic to {learn a latent space from a data which can represent the input space}. It can not only grasp the hidden structure of the data, but also generate low dimensional features by nonlinear mapping. A large amount of manifold learning methods have already been proposed, such as the Isometric Feature Mapping (ISOMAP) \cite{isomap, Takwalkar}, Laplacian Eigenmaps \cite{laplacian,Takwalkar}, Local linear Embedding (LLE) \cite{lle}, Semidefinite Embedding \cite{sde}, Manifold Discriminant Analysis \cite{manifold_distance,manifold_manifold_distance}, RSR-ML \cite{Harandi2014}. With the development of the deep learning, some works have incorporated the manifold in the deep models \cite{Aziere2019,nature,lu2015,Iscen2018}. Zhu \textit{et al.} \cite{nature} develops the automated transform by manifold approximation (AUTOMAP) which learns a near-optimal reconstruction mapping through manifold learning. Lu \textit{et al.} \cite{lu2015} and Aziere \textit{et al.} \cite{Aziere2019} mainly apply the manifold learning in deep ensemble and consider the manifold similarity relationships between different CNNs. Iscen \textit{et al.} \cite{Iscen2018} utilizes the manifolds to implement the metric learning without labels.

{However, these prior works} are mainly applied in natural image processing tasks, such as face recognition \cite{Harandi2014}, natural image classification \cite{manifold_manifold_distance}, image retrieval \cite{Aziere2019}.
Only few works, such as \cite{li2010}, \cite{wang2016}, \cite{duan2020}, focus on the hyperspectral image classification task. Among these works, Ma \textit{et al.} \cite{li2010} combines the local manifold learning with the $k$-nearest-neighbor classifier. Wang \textit{et al.} \cite{wang2016} uses the manifold ranking for salient band selection.
Duan \textit{et al.} \cite{duan2020} combines the manifold structure and sparse relationship for discriminative features.
{Furthermore, few of these works consider the manifold structure of hyperspectral image in the deep learning process.}
Faced with the current task, this work tries to develop a novel training loss with the {special manifold structure of each class in } hyperspectral image which can promote the learned deep model to capture the data intrinsic {manifold structure} and further improve the representational ability of the deep model. In the following, we will introduce the developed deep manifold embedding in detail.

\begin{figure*}[t]
\centering
 \includegraphics[width=0.99\linewidth]{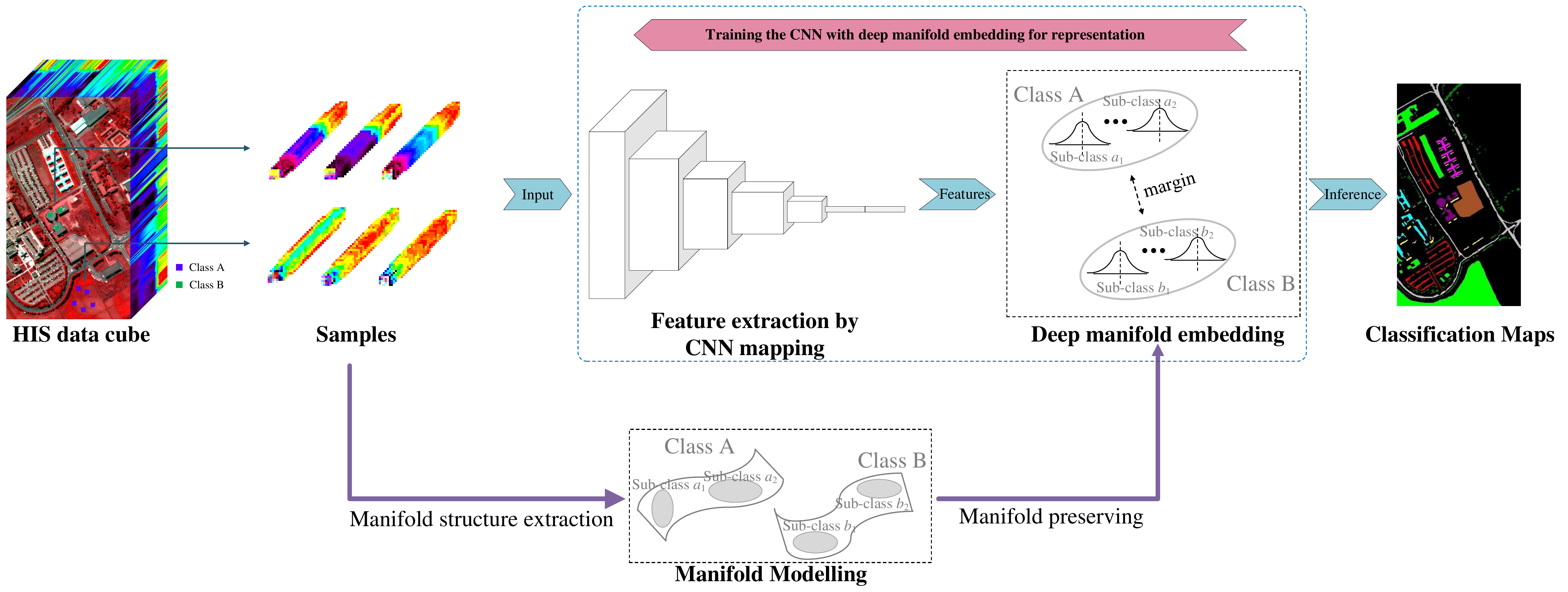}
   \caption{Flowchart of the proposed deep manifold embedding for hyperspectral image classification.}
\label{fig:flowchart}
\end{figure*}

\section{Manifold Embedding in Deep Learning}\label{sec:manifold_embedding}

Denote $X=\{{\bf x}_1,{\bf x}_2,\cdots,{\bf x}_n\}$ as the training samples of the hyperspectral image and $y_i$ is the corresponding class label of ${\bf x}_i$ where $n$ defines the number of the training samples. $y_i\in \Gamma=\{1,2,\cdots,{\Lambda}\}$ where $\Gamma$ stands for the set of class labels and $\Lambda$ represents the number of the class of the image.

The flowchart of the proposed deep manifold embedding {method} for hyperspectral image classification is shown in Fig. \ref{fig:flowchart}. The main process can be divided into three parts: manifold modelling, the construction of the deep manifold embedding, and the training process of the CNN with the developed training loss.

\subsection{Manifold Modelling}\label{subsec:manifold_modelling}

Let $C_{s}=\{{\bf x}_{h_1},{\bf x}_{h_2},\cdots,{\bf x}_{h_{n_{s}}}\}$ denote the set of samples from the $s-$th class among the whole training samples, where $n_{s}$ is the number of samples from the $s-$th class and $h_i$ describes the subscript of the training sample.

Following the law of manifold distribution,
samples of each class from the hyperspectral image are assumed to satisfy a certain nonlinear manifold.
Then, according to the law of cluster distribution on nonlinear manifold, each class can be divided into several sub-classes and each sub-class is supposed to follow a certain probability distribution.

\begin{algorithm}[t]
\renewcommand{\algorithmicrequire}{\textbf{Input:}}
\renewcommand{\algorithmicensure}{\textbf{Output:}}
\caption{Manifold Modelling via Hierarchical Clustering}\label{algorithm:01}
\begin{algorithmic}[1]
\REQUIRE ${X={\{{\bf x}_1,{\bf x}_2,\cdots,{\bf x}_n\}}}$, $C_{s}=\{{\bf x}_{h_1}, {\bf x}_{h_2}, \cdots, {\bf x}_{h_{n_{s}}}\}$ $ (s=1,2,\cdots,\Lambda)$, $k$, $b$
\ENSURE $T_1^{(s)},T_2^{(s)},\cdots,T_k^{(s)}$,$s=1,2,\cdots,\Lambda$
\FOR{$s=1,2,\cdots,\Lambda$}
\STATE Construct the undirected graph over $s-$th class with the node of $C_{s}$.
\STATE Compute the weights of edges on the graph using Eq. \ref{eq:02}.
\STATE Compute the distance matrix $S$ over the manifold using Eq. \ref{eq:03} through Dijkstra algorithm.
\WHILE{obtain the $k$ sub-classes}
\STATE Combine the nearest two set points in distance matrix as a new set.
\STATE Update the distance matrix $S$ with the newly established set.
\ENDWHILE
\ENDFOR
\STATE {\bf return} $T_1^{(s)},T_2^{(s)},\cdots,T_k^{(s)}$,$s=1,2,\cdots,\Lambda$
\end{algorithmic}
\end{algorithm}

Given the $s-$th class in the image.
To {divide} the samples of each class into different sub-classes, all the samples of each class {are} used to formulate an undirected graph. Let $G_{s}=(V_{s}, E_{s})$ denote the graph over the $s-$th class,  where $V_{s}=\{{\bf x}_{h_1},{\bf x}_{h_2},\cdots,{\bf x}_{h_{n_{s}}}\}$ is the set of  {nodes and} $E=\binom{V_{s}}{2}$ is the set of edges in the graph. 

As shown in Fig. \ref{fig:flowchart}, {the closer samples on the manifold have higher probability to belong to the same sub-class.}
Therefore, the distance between the sample ${\bf x}_{h_i}$ and its $b$ nearest neighbors $N_b({\bf x}_{h_i})$ is assumed to distribute on a certain linear manifold and can be calculated under the Euclidean distance,
\begin{equation}\label{eq:01}
  D({\bf x}_{h_i},{\bf x}_{h_j})=\|{\bf x}_{h_i}-{\bf x}_{h_j}\|, \text{if} \ {\bf x}_{h_j}\in N_b({\bf x}_{h_i})
\end{equation}

Then, the weights of the edges on the undirected graph {of} the $s-$th class are defined as follows:
\begin{equation}\label{eq:02}
  W({{\bf x}_{h_i},{\bf x}_{h_j}})=\left\{
 \begin{aligned}
D({\bf x}_{h_i},{\bf x}_{h_j}),\ &\text{if} \ {\bf x}_{h_j}\in N_b({\bf x}_{h_i}) \\
 \infty\ \ \ \ \ \ ,\ &\text{if}  \ {\bf x}_{h_j}\notin N_b({\bf x}_{h_i})
 \end{aligned}
  \right.
\end{equation}

{Generally, the geodesic distance \cite{geodesic} can be used to measure the distance between different samples on the manifold \cite{isomap}}.
The geodesic distance on the manifold can be transformed {as} the shortest path on the graph $G_{s}$. Then, the distance between the sample ${\bf x}_{h_p}$ and ${\bf x}_{h_q}$ on the manifold can be calculated by
\begin{equation}\label{eq:03}
  S({\bf x}_{h_p},{\bf x}_{h_q})=\min\limits_l\min\limits_{{\bf x}_{\gamma_1},\cdots,{\bf x}_{\gamma_l}\in C_{s} }\sum_{i=2}^{l}\|{\bf x}_{\gamma_i}-{\bf x}_{\gamma_{i-1}}\|
\end{equation}
where ${\bf x}_{\gamma_1}={\bf x}_{h_p},{\bf x}_{\gamma_l}={\bf x}_{h_q}, {\bf x}_{\gamma_i}\in N_b({\bf x}_{\gamma_{i-1}}) (i=2,\cdots,l)$.

This work uses the Dijkstra algorithm \cite{dijkstra} to solve the optimization in Eq. \ref{eq:03}. Then, the distance matrix $S$ over the data manifold of the $s-$th class can be formulated by the pairwise distance $S({\bf x}_{h_p},{\bf x}_{h_q})$ between different samples.

Here, for each class, we divide the whole training  samples of the class into $k$ sub-classes.
Denote $T_1,T_2,\cdots,T_k$ as the $k$ sub-classes of the $s-$th class. {Then,} these $k$ sub-classes {can be} constructed {by solving} the following optimization:
\begin{equation}\label{eq:04}
\min\limits_{T_1,T_2,\cdots,T_k}\max\limits_{i\in \{1,2,\cdots,k\}}\max\limits_{{\bf x}_{m},{\bf x}_{n}\in T_i} S({\bf x}_{m},{\bf x}_{n})
\end{equation}
Under the optimization in Eq. \ref{eq:04}, we can obtain the $k$ sub-classes {which have} the smallest geodesic distances between the samples in each sub-class.
Hierarchical clustering can be used to solve the optimization.
The whole procedure is outlined in Algorithm \ref{algorithm:01}.

\subsection{Deep Manifold Embedding for Training Loss}

This work selects the CNN model as the features extracted model for hyperspectral image.
Denote $\varphi({\bf x}_i)$ as the extracted features of sample ${\bf x}_i$ from the CNN model.
Then, the obtained features can be looked as the global low dimensional coordinates under the nonlinear CNN mapping. Besides, as Fig. \ref{fig:flowchart} shows, the deep manifold embedding constructs the global low dimensional coordinates to preserve the estimated distance on the manifold.


As processed in former subsection, suppose $T_1^{(s)},T_2^{(s)},\cdots,T_k^{(s)}$ as the $k$ sub-classes from the $s-$th class. Given $T_e^{(s)}=\{{\bf x}_{z_1},{\bf x}_{z_2},\cdots,{\bf x}_{z_{n_{T_e}}}\}(1\leq e\leq k)$ where $z_i(i=1,\cdots,n_{T_e})$ is the subscript of the training sample and $n_{T_e^{(s)}}$ is the number of samples in the sub-class $T_e^{(s)}$. If not specified, in the following, we use $T_e$ to represent the $T_e^{(s)}$.  Then, $R_e=\{\varphi({\bf x}_{z_1}),\varphi({\bf x}_{z_2}),\cdots,\varphi({\bf x}_{z_{n_{T_e}}})\}$ is the set of the learned features.

Following the law of cluster distribution, a sub-class follows a certain probability distribution {on} the manifold.
It can be transformed to the one that $\forall \varphi({\bf x}_{z_i})\in R_e$, $\varphi({\bf x}_{z_i})$ follows the {distribution} constructed by all the other features in $R_e$ under a certain degree of confidence.

Therefore, given $\varphi({\bf x}_{z_o})\in R_e$, suppose all the other features in $R_e$ follow the multi-variant Gaussian distribution $R_e\sim N_p(\mu,\Sigma)$ where $p$ is the dimension of the learned features. Then,
\begin{equation}\label{eq:05}
  {{(R_e-\mu)^T\Sigma^{-1}(R_e-\mu)}{} \sim \chi^2(p)}
\end{equation}
{where $\chi^2(p)$ represents the $\chi^2-$distribution with $p$ degrees of freedom which is the distribution of a sum of the squares of $p$ independent standard normal random variables.}
Under the confidence $\alpha$, when
\begin{equation}\label{eq:06}
  {(\varphi({\bf x}_{z_o})-\mu)^T\Sigma^{-1}(\varphi({\bf x}_{z_o})-\mu)}{}<\chi^2_p(\alpha),
\end{equation}
{where $\chi^2_p(\alpha)$ means the upper bound of the confidence interval for $\chi^2(p)$-distribution under confidence $\alpha$.}
$\varphi({\bf x}_{z_o})$ can be seen as the sample from distribution $R_e\sim N_p(\mu,\Sigma)$.
For simplicity, we assume that different dimensions of the learned features are independent and have the same variance \cite{gong_statistical}, namely the covariance $\Sigma=\sigma^2I_0$ where $I_0$ represents the identity matrix.
Besides, the unbiased estimation of the mean value $\mu$ can be {formulated} as
\begin{equation}\label{eq:07}
  \hat{\mu}=\frac{1}{n_{T_e}-1}\sum\limits_{z_i\neq z_o}\varphi({\bf x}_{z_i}).
\end{equation}
Then, the penalization from ${T_e}$ can be {calculated} by
\begin{equation}\label{eq:08}
  L_{T_e}=\eta\sum_{o=1}^{n_{T_e}}(\sum_{i=1}^{n_{T_e}}(\varphi({\bf x}_{z_o})-\varphi({\bf x}_{z_i})))^T(\sum_{i=1}^{n_{T_e}}(\varphi({\bf x}_{z_o})-\varphi({\bf x}_{z_i})))
\end{equation}
where $\eta$ is the constant term. Since
\begin{equation}\label{eq:09}
\begin{aligned}
  (\sum_{i=1}^{n_{T_e}}(\varphi({\bf x}_{z_o}&)-\varphi({\bf x}_{z_i})))^T(\sum_{i=1}^{n_{T_e}}(\varphi({\bf x}_{z_o})-\varphi({\bf x}_{z_i})))\leq \\
  &2\sum_{i=1}^{n_{T_e}}(\varphi({\bf x}_{z_o})-\varphi({\bf x}_{z_i}))^T(\varphi({\bf x}_{z_o})-\varphi({\bf x}_{z_i}))
\end{aligned}
\end{equation}
Ignore the constant term and we can use the following penalization to optimize Eq. \ref{eq:08},
\begin{equation}\label{eq:10}
  L_{T_e}=\sum_{o=1}^{n_{T_e}}\sum_{i=1}^{n_{T_e}}(\varphi({\bf x}_{z_o})-\varphi({\bf x}_{z_i}))^T(\varphi({\bf x}_{z_o})-\varphi({\bf x}_{z_i}))
\end{equation}

Then, the loss for manifold embedding in the deep learning process can be written as
\begin{equation}\label{eq:11}
  L_0=\sum_{s=1}^{\Lambda}\sum_{e=1}^{k}L_{T_e^{(s)}}.
\end{equation}

To further improve the performance {of} manifold embedding, we introduce the diversity-promoting term to enlarge the distance between the sub-classes from different classes.
{The distance} can be {measured} by the {set-to-set distance}.
This work {chooses} the Hausdorff distance which is the maximum distance of a set to the nearest point in the other set \cite{hausdorff} to measure the distance between different sub-classes since the measurement considers {not only} the whole shape of the data set {but also} the position of the samples in the set.

Suppose $T_m^{(s)}$ as the $m$ sub-class from $s-$th class and $T_n^{(t)}$ as the $n$ sub-class from $t-$th class, then the Hausdorff distance between the two sub-classes can be calculated by
\begin{equation}\label{eq:12}
  D_H(T_m^{(s)},T_n^{(t)})=\max\limits_{{\bf x}_p\in T_m^{(s)}}\min \limits_{{\bf x}_q\in T_n^{(t)}}\|\varphi({\bf x}_p)-\varphi({\bf x}_q)\|^2
\end{equation}

Then, the diversity-promoting term \cite{gong_diversity} can be formulated as
\begin{equation}\label{eq:13}
{L_d=\sum_{s=1}^{\Lambda}\sum_{s\neq t}^{\Lambda}\sum_{m=1}^{k}\sum_{n=1}^{k}[\Delta-D_H(T_m^{(s)},T_n^{(t)})],}
\end{equation}
where $\Delta$ is a positive value which represents the margin.

Based on Eq. \ref{eq:11} and \ref{eq:13}, the final loss for the proposed DMEM can be written as
\begin{equation}\label{eq:14}
  L=L_0+\beta L_d
\end{equation}
where $\beta$ stands for the tradeoff parameter.

\subsection{Training}

Just as general deep learning methods, stochastic gradient descent (SGD) methods and back propagation (BP) \cite{backpropagation} are used for the training process of the developed deep manifold embedding \footnote{{The code for the implementation of the proposed method will be released soon at \url{http://github.com/shendu-sw/deep-manifold-embedding}.}}. The key process is to calculate the derivation of the loss $L$ with respect to (w.r.t.) the features $\varphi({\bf x}_i)$.

Based on the chain rule, gradients of $L$ w.r.t. $\varphi({\bf x}_i)$ can be calculated as
\begin{equation}\label{eq:15}
  \frac{\partial L}{\partial \varphi({\bf x}_i)}=\frac{\partial L_0}{\partial \varphi({\bf x}_i)}+\beta \frac{\partial L_d}{\partial \varphi({\bf x}_i)}
\end{equation}

Then, we have
\begin{equation}\label{eq:16}
  \frac{\partial L_0}{\partial \varphi({\bf x}_i)}=\sum_{s=1}^{\Lambda}\sum_{e=1}^{k}I({\bf x}_i\in T_e^{s})\sum\limits_{{\bf x}_p\in T_e^{s}}(\varphi({\bf x}_i)-\varphi({\bf x}_p))
\end{equation}
where $I(\cdot)$ represents the indicative function.
\begin{equation}\label{eq:17}
\begin{aligned}
  \frac{\partial L_d}{\partial \varphi({\bf x}_i)}= &
  \sum_{t=1}^{\Lambda}\sum_{n=1}^{k}\sum\limits_{{\bf x}_q\in T_n^{(t)}}(\varphi({\bf x}_q)-\varphi({\bf x}_i))I((\|\varphi({\bf x}_q)-\\ &\varphi({\bf x}_i)\|^2 =D_H(T_m^{(s)}, T_n^{(t)})) \cap  ({\bf x}_i\in T_n^{(t)} )),
\end{aligned}
\end{equation}
{We summarize the computation of loss functions and gradients in Algorithm \ref{algorithm:02}.
It should be noted that the whole CNN is trained under joint supervisory of softmax loss and our DMEM loss. Then, the extracted features from the CNN are tranferred to the Softmax classifier for classification.}

\begin{algorithm}[t]
\renewcommand{\algorithmicrequire}{\textbf{Input:}}
\renewcommand{\algorithmicensure}{\textbf{Output:}}
\caption{Calculate Gradient for DMEM}\label{algorithm:02}
\begin{algorithmic}[1]
\REQUIRE Features $\{\varphi({\bf x}_1), \varphi({\bf x}_2), \cdots, \varphi({\bf x}_n)\}$, $C=\{C_{1}, $ $C_{2},\cdots, C_{{\Lambda}}\}$, sub-classes $\{T_1^{(s)}, T_2^{(s)}, \cdots, T_k^{(s)}\}(s=1,2,\cdots,\Lambda)$, hyperparameter $\Delta, \beta$.
\ENSURE $L_0, L_d, \frac{\partial L_0}{\partial \varphi({\bf x}_i)}, \frac{\partial L_d}{\partial \varphi({\bf x}_i)}$
\STATE Compute the loss for deep manifold embedding in each sub-class in the mini-batch using Eq. \ref{eq:10}.
\STATE Compute the loss $L_0=\sum_{s=1}^{\Lambda}\sum_{e=1}^{k}L(T_e^{(s)})$.
\STATE Compute the Hausdroff distance $D_H(T_m^{(s)},T_n^{(t)})$ between sub-classes from different classes using Eq. \ref{eq:12}.
\STATE Compute the diversity-promoting term by $L_d=$ $\sum_{s=1}^{\Lambda}\sum_{t=1}^{\Lambda}\sum_{m=1}^{k}\sum_{n=1}^{k}[\Delta-D_H(T_m^{(s)},T_n^{(t)})]$.
\STATE Compute $\frac{\partial L_0}{\partial \varphi({\bf x}_i)}$ using Eq. \ref{eq:16}.
\STATE Compute $\frac{\partial L_d}{\partial \varphi({\bf x}_i)}$ using Eq. \ref{eq:17}.
\STATE {\bf return} $L_0, L_d, \frac{\partial L_0}{\partial \varphi({\bf x}_i)}, \frac{\partial L_d}{\partial \varphi({\bf x}_i)}$.
\end{algorithmic}
\end{algorithm}

\section{Experimental Results}\label{sec:experiment}

In this section, intensive experiments are conducted to prove the effectiveness of the proposed method. First, the datasets used in this work are introduced. Then, the experimental setups are detailed and the experimental results are shown and analyzed.

\subsection{Datasets}

To further validate the effectiveness of the proposed method, this work conducts experiments over four real-world hyperspectral images, namely the Pavia University \cite{data}, the Salinas Scene \cite{data}, {the Houston 2013 \cite{houston2013}  and the Houston 2018 data \cite{houston2018}}.
\begin{enumerate}
  \item The Pavia University was acquired by the reflective optics system imaging spectrometer (ROSIS-3) sensor during a flight campaign over Pavia, Northern Italy. The image consists of $610\times 340$ pixels with a geometric resolution of 1.3 m/pixels. A total of 42,776 labelled samples divided into 9 land cover objects are used for experiments and each sample is with 103 spectral bands ranging from 0.43 to 0.86 $\mu m$. 
  \item The Salinas Scene was also collected by the 224-band AVIRIS sensor with a spectral coverage from 0.4 to 2.5 $\mu m$ but over Salinas Valley, California. The image size is $512 \times 217$ with a spatial resolution of 3.7 m/pixel. 20 water absorption bands are discarded. 16 classes of interest, including vegetables, bare soils, and vineyard fields with a total of 54,129 labelled samples are chosen for experiments. 
  \item {The Houston 2013 data was published as part of the 2013 IEEE Geoscience and Remote Sensing Society data fusion contest with a spectral coverage ranging from 0.38 to 1.05 $\mu m$. The image size is $349 \times 1905$ with a spatial resolution of 2.5 m/pixel. The data contains 144 spectral bands and 15 classes of interes with a total of 15,011 labelled samples are chosen for experiments.}
  \item {The Houston 2018 data is similar to Houston 2013 data which has a larger spatial size (i.e., 601 $\times$ 2384) but less spectral bands (i.e., 48). 20 different land-cover classes with a total of 504,712 labelled samples are chosen for experiments.}
\end{enumerate}

\subsection{Experimental Setups}

There are four parameters in the experiments to be determined, namely the {balance parameter $\beta$ between the optimization term and the diversity-promoting term}, {the balance parameter $\lambda$} between the manifold embedding term and the softmax loss, the number of sub-classes $k$, {and the} number of the neighbors $b$. The first two are empirically set as $\lambda=0.0001, {\beta=0.0001}$. As for $k$ and $b$, a lot of experiments have been done to choose the best parameters. We set the two variables as different values and then check their performance under various $k$ and $b$.

\begin{figure*}[t]
\centering
 \includegraphics[width=0.98\linewidth]{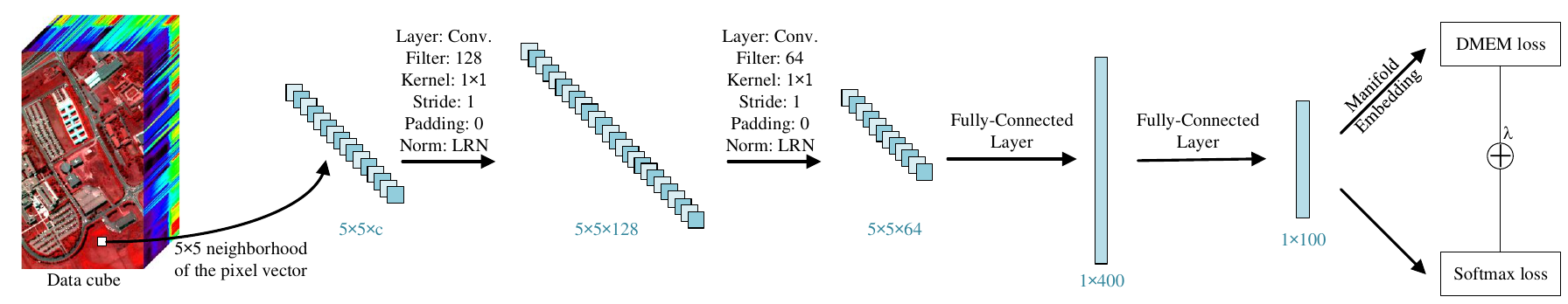}
   \caption{{Architecture of CNN model for hyperspectral image classification. In the figure, $c$ denotes the channel bands of the image. }}
\label{fig:cnn_model}
\end{figure*}

Caffe \cite{caffe} is chosen as the deep learning framework to implement the developed method for hyperspectral image classification.
{This work adopts the simple CNN architecture as Fig. \ref{fig:cnn_model} shows to provide the nonlinear mapping for the low dimensional features of the data manifold.
As the figure shows, the input of the CNN is $5\times5\times c$ where $c$ denotes the channel bands of the image. Two convolutional layers with $1\times1$ kernel are adopted to extract the spectral information and the following two fully connected layers are used to extract the global spatial information.}
{The number of parameters in the proposed CNN model is about 2.70M (e.g. 2.68M, 2.73M, 2.70M, 2.65M for Pavia University, Salinas Scene data, Houston2013, and Houston2018, respectively).
It should be noted that about 97\% of the parameters are from the fully-connected layer.
}

\begin{table*}
\scriptsize
  \centering
  \caption{{Number of training and testing samples.}}
  \begin{tabular}{|c | c | c| c | c| c| c| c |c| c| c  |c| c|}
  \hline
  \multirow{2}{*}{\textbf{ID}}  & \multicolumn{3}{c|}{\bf Pavia University} & \multicolumn{3}{c|}{\bf Salinas Scene}  & \multicolumn{3}{c|}{\bf Houston2013} & \multicolumn{3}{c|}{\bf Houston2018} \\
\cline{2-13}
  & Class Name & Train & Test & Class Name & Train & Test & Class Name & Train & Test & Class Name & Train & Test \\
  \hline\hline
  C1 & Asphalt &200 & 6431 & Brocoli-green-weeds-1 & 200 & 1809 & grass-healthy & 198 & 1053 & grass-healthy& 1458 & 8341\\
  C2 & meadows&200 & 18449 & Brocoli-green-weeds-2&200 & 3526& grass-stressed& 190 & 1064& grass stressed & 4316 & 28186\\
  C3 &gravel& 200 & 1899 & Fallow&200 & 1776& grass-synthetic & 192 & 505& artificial turf & 331 & 353\\
  C4 & trees & 200 & 2864& Fallow-rough-plow&200 & 1194& tree& 188 & 1056& evergreen trees & 2005 & 11583\\
  C5 &metal sheet& 200 & 1145&Fallow-smooth& 200 & 2478&soil&  186 & 1056& deciduous trees &676 & 4372\\
  C6 &bare soil& 200 & 4829& stubble&200 & 3759&water&  182 & 143& bare earth& 1757 & 2759\\
  C7 &bitumen& 200 & 1130& celery&200 & 3379& residential & 196 & 1072& water  &147 & 119\\
  C8 &brick& 200 & 3482& grapes-untrained&200 & 11071& commercial&  191 & 1053& residential buildings  & 3809 & 35953\\
  C9 &shadow& 200 & 747& soil-vinyard-develop&200 & 6003&road&  193 & 1059& non-residential buildings & 2789 & 220895\\
  C10 && & &corn-senesced-green-weeds&200 & 3078&highway& 191 & 1036 & roads& 3188 & 42622 \\
  C11 && & &lettuce-romaine-4wk&200 & 868 &railway & 181 & 1054 & sidewalks & 2699 & 31303\\
  C12 && & &lettuce-romaine-5wk&200 & 1727 &parking-lot1& 192 & 1041 & crosswalks & 225 & 1291 \\
  C13 && & &lettuce-romaine-6wk &200 & 716 &parking-lot2& 184 & 285 & major thoroughfares& 5193&41165 \\
  C14 && & &lettuce-romaine-7wk & 200& 870 &tennis-court& 181 & 247 & highways & 700 &9149 \\
  C15 && & &vinyard-untrained&200 & 7068 &running-track& 187 & 473 & railways & 1224&5713 \\
  C16 && & &vinyard-vertical-trellis&200 & 1607 &&  &  & paved parking lots &1179 &10296 \\
  C17 && & && &  & &&   & unpaved parking lots & 127& 22\\
  C18 && & && &  & & &  & cars & 848& 5730 \\
  C19 && & && &  & & &  &trains & 493 &4872 \\
  C20 && & && &  & & &  & stadium seats&1313 &5511 \\
\hline
Total & & 1800 & 40976 &  & 3200 & 50929 & & 2832 & 12197 & & 34477 & 470235\\
  \hline
  \end{tabular}
  \label{table:tr_test}
\end{table*}

The learning rate, epoch iteration, training batch are set to 0.001, 60000, 84, respectively.
Just as Fig. \ref{fig:cnn_model}, this work takes advantage of the $5\times 5$ neighbors to extract both the spatial and the spectral information from the image.
{In the experiments, we choose training samples randomly over Pavia University and Salinas scene data. {Besides, for Houston2013 data and Houston2018 data, the training set are generated by the
Image Analysis and Data Fusion Technical Committee of the
IEEE Geoscience and Remote Sensing Society (GRSS) and
the University of Houston for the 2013 data fusion contest and 2018 data fusion contest separately.} The detailed numbers of training and testing samples for the experiments have been described in Table \ref{table:tr_test}.}
To objectively evaluate the classification performance, metrics of the overall accuracy (OA), average accuracy (AA), and the Kappa coefficient are adopted. All the results come from the average value and standard deviation of ten runs of training and testing.


\subsection{{Results over Pavia University and Salinas Scene Dataset}}

\subsubsection{General Performance}
At first, we present the general performance of the developed manifold embedding for hyperspectral image classification. In this set of experiments, the number of sub-classes $k$ is set to 5, the number of neighbors $b$ is set to 5.
Very common machine with a 3.6-GHz Intel Core i7 CPU, 64-GB memory and NVIDIA GeForce GTX 1080 GPU was used to test the performance of the proposed method. The proposed method took about 2196s over Pavia University data, and 2965s over Salinas scene data. It should be noted that the developed manifold embedding is implemented through CPU and the computational performance can be remarkably improved by modifying the codes to run on the GPUs.

{Table \ref{table:pavia} and \ref{table:salinas} show the general performance over the Pavia University and salinas scene data}, respectively. These tables show the classification accuracies of each class and the OA, AA as well as the Kappa by SVM-POLY, the CNN trained with softmax loss and the CNN trained with the proposed method.
From these tables, we can easily get that the CNN model provides a more discriminative representation of the hyperspectral image than other handcrafted features.
{Furthermore, we can find that the proposed DMEM method can improve the performance of the vanilla CNN.}
Over the Pavia University data, the CNN model with the manifold embedding can obtain an accuracy of ${99.52\%\pm 0.10\%}$ which is higher than $98.61\%\pm 0.35\%$ by the CNN with softmax loss only. Over the Salinas scene data, the proposed method which can also achieve ${97.80\%\pm 0.21\%}$ outperforms the CNN with general softmax loss.


To further validate the effectiveness of the developed method, this work uses the McNemar's test \cite{mcnemar}, which is based on the standardized normal test statistics, for deeply comparisons in the statistic sense. The statistic can be computed by
\begin{equation}\label{eq:09_mcnemar}
  F_{ij}=\frac{f_{ij}-f_{ji}}{\sqrt{f_{ij}+f_{ji}}}
\end{equation}
where $f_{ij}$ describes the number of correctly classified samples by the $i$th method but wrongly by the $j$th method. Therefore, $F_{ij}$ measures the pairwise statistical significance between the $i$th and $j$th methods. At the widely used $95\%$ level of confidence, the difference of accuracies between different methods is statistically significant if $|F_{ij}|>1.96$.

\begin{table}[t]
\begin{center}
\caption{Classification accuracies ($Mean\pm SD$) (OA, AA, and Kappa) of different methods achieved on the Pavia University data. The results from CNN is trained with the Softmax Loss. ${|F_{ij}|}$ represents the value of McNemar's test.}
\label{table:pavia}
\begin{tabular}{|c | c | c c c|}
\hline
\multicolumn{2}{|c|}{\bf Methods}     &  {\bf SVM-POLY} &  {\bf CNN} &  {\bf Proposed Method} \\
\hline\hline
\multirow{9}{*}{\rotatebox{90}{\tabincell{c}{\textbf{Classification} \\ \textbf{Accuracies (\%)}}}}          & C1   &  $83.01\pm 1.30$  &  $98.50\pm 0.49$ &  $\mathbf{99.68\pm 0.17}$\\
                                                                                                             & C2   &  $86.61\pm 1.80$  &  $99.02\pm 0.59$ &  $\mathbf{99.75\pm 0.12}$\\
                                                                                                             & C3   &  $85.96\pm 1.04$  &  $95.92\pm 2.84$ &  $\mathbf{97.33\pm 1.56}$\\
                                                                                                             & C4   &  $96.36\pm 0.92$  &  $98.78\pm 0.55$ &  $\mathbf{99.21\pm 0.47}$\\
                                                                                                             & C5   &  $99.62\pm 0.18$  &  $100.0\pm 0.00$ &  $\mathbf{100.0\pm 0.00}$\\
                                                                                                             & C6   &  $90.96\pm 1.57$  &  $99.36\pm 1.00$ &  $\mathbf{99.68\pm 0.34}$\\
                                                                                                             & C7   &  $93.92\pm 0.80$  &  $99.56\pm 0.36$ &  $\mathbf{99.81\pm 0.17}$\\
                                                                                                             & C8   &  $87.27\pm 1.56$  &  $95.90\pm 3.31$ &  $\mathbf{98.87\pm 0.50}$\\
                                                                                                             & C9   &  $99.93\pm 0.13$  &  $100.0\pm 0.00$ &  $\mathbf{100.0\pm 0.00}$\\
 \hline
 \multicolumn{2}{|c|}{{\bf OA}  (\%)}      &  $88.07\pm 0.82$ &  $98.61\pm 0.35$  &  $\mathbf{99.52\pm 0.10}$\\
 \hline
 \multicolumn{2}{|c|}{{\bf AA}  (\%)}      &  $91.52\pm 0.26$ &  $98.56\pm 0.36$  &  $\mathbf{99.37\pm 0.17}$\\
 \hline
 \multicolumn{2}{|c|}{{\bf KAPPA} (\%)}    &  $84.35\pm 1.01$ &  $98.14\pm 0.47$  &  $\mathbf{99.35\pm 0.14}$\\
\hline
\multicolumn{2}{|c|}{{${|F_{ij}|}$}}    &  $63.95$ &  $20.80$  &  $-$\\
\hline
\end{tabular}
\end{center}
\end{table}

From these tables, it can also be noted that when compared the proposed method with the CNN trained by general softmax loss, the Mcnemar's test value $|F_{ij}|$  achieves 20.80 and 12.67 over Pavia University and salinas scene data, respectively. This indicates that the improvement of the developed deep manifold embedding on the performance of CNN is statistically significant.

\begin{table}[t]
\begin{center}
\caption{Classification accuracies ($Mean\pm SD$) (OA, AA, and Kappa) of different methods achieved on the Salinas Scene data. }
\label{table:salinas}
\begin{tabular}{|c | c | c c c|}
\hline
\multicolumn{2}{|c|}{\bf Methods}     &  {\bf SVM-POLY} &  {\bf CNN} &  {\bf Proposed Method} \\
\hline\hline
\multirow{16}{*}{\rotatebox{90}{\tabincell{c}{\textbf{Classification} \\ \textbf{Accuracies (\%)}}}}          & C1   &  $99.65\pm 0.15$  &  $99.89\pm 0.15$ &  $\mathbf{99.98\pm 0.07}$\\
                                                                                                             & C2   &  $99.87\pm 0.09$  &  $99.91\pm 0.09$ &  $\mathbf{99.91\pm 0.11}$\\
                                                                                                             & C3   &  $99.61\pm 0.17$  &  $99.98\pm 0.07$ &  $\mathbf{100.0\pm 0.00}$\\
                                                                                                             & C4   &  $99.53\pm 0.16$  &  $99.69\pm 0.27$ &  $\mathbf{99.77\pm 0.16}$\\
                                                                                                             & C5   &  $98.37\pm 0.51$  &  $99.53\pm 0.35$ &  $\mathbf{99.63\pm 0.47}$\\
                                                                                                             & C6   &  $99.77\pm 0.24$  &  $\mathbf{100.0\pm 0.00}$ &  ${99.99\pm 0.04}$\\
                                                                                                             & C7   &  $99.62\pm 0.23$  &  $99.86\pm 0.15$ &  $\mathbf{99.88\pm 0.09}$\\
                                                                                                             & C8   &  $79.13\pm 2.73$  &  $92.76\pm 1.85$ &  $\mathbf{94.86\pm 1.00}$\\
                                                                                                             & C9   &  $99.47\pm 0.39$  &  $99.90\pm 0.10$ &  $\mathbf{99.94\pm 0.08}$\\
                                                                                                             & C10   &  $93.25\pm 0.70$  &  $98.54\pm 0.66$ &  $\mathbf{99.19\pm 0.44}$\\
                                                                                                             & C11   &  $98.65\pm 0.80$  &  $99.38\pm 0.49$ &  $\mathbf{99.78\pm 0.21}$\\
                                                                                                             & C12   &  $99.93\pm 0.05$  &  $99.91\pm 0.15$ &  $\mathbf{100.0\pm 0.00}$\\
                                                                                                             & C13   &  $99.05\pm 0.45$  &  $99.92\pm 0.19$ &  $\mathbf{99.97\pm 0.09}$\\
                                                                                                             & C14   &  $97.06\pm 0.74$  &  $99.75\pm 0.37$ &  $\mathbf{99.90\pm 0.15}$\\
                                                                                                             & C15   &  $73.77\pm 1.96$  &  $91.10\pm 2.70$ &  $\mathbf{92.99\pm 1.22}$\\
                                                                                                             & C16   &  $99.09\pm 0.33$  &  $99.52\pm 0.48$ &  $\mathbf{99.63\pm 0.35}$\\
 \hline
 \multicolumn{2}{|c|}{{\bf OA}  (\%)}      &  $91.07\pm 0.42$ &  $97.01\pm 0.22$  &  $\mathbf{97.80\pm 0.21}$\\
 \hline
 \multicolumn{2}{|c|}{{\bf AA}  (\%)}      &  $95.99\pm 0.13$ &  $98.73\pm 0.12$  &  $\mathbf{99.09\pm 0.07}$\\
 \hline
 \multicolumn{2}{|c|}{{\bf KAPPA} (\%)}    &  $90.01\pm 0.46$ &  $96.65\pm 0.25$  &  $\mathbf{97.54\pm 0.23}$\\
\hline
\multicolumn{2}{|c|}{{${|F_{ij}|}$}}    &  $52.04$ &  $12.67$  &  $-$\\
\hline
\end{tabular}
\end{center}
\end{table}

\subsubsection{Effects of Different Number of Training Samples}

Since the number of training samples can significantly affect the construction of the data manifold, this part will further validate the performance of the developed deep manifold embedding under different number of training samples. For the Pavia University and the Salinas Scene data, the number of training samples per class is selected from $\{10, 20, 40, 80, 120, 160, 200\}$. In this set of experiments, the number of the sub-classes $k$ and the neighbors $b$ is set to 5, 5, respectively.

\begin{figure}[t]
\centering
 \subfigure[]{\label{fig:number_pavia}\includegraphics[width=0.48\linewidth]{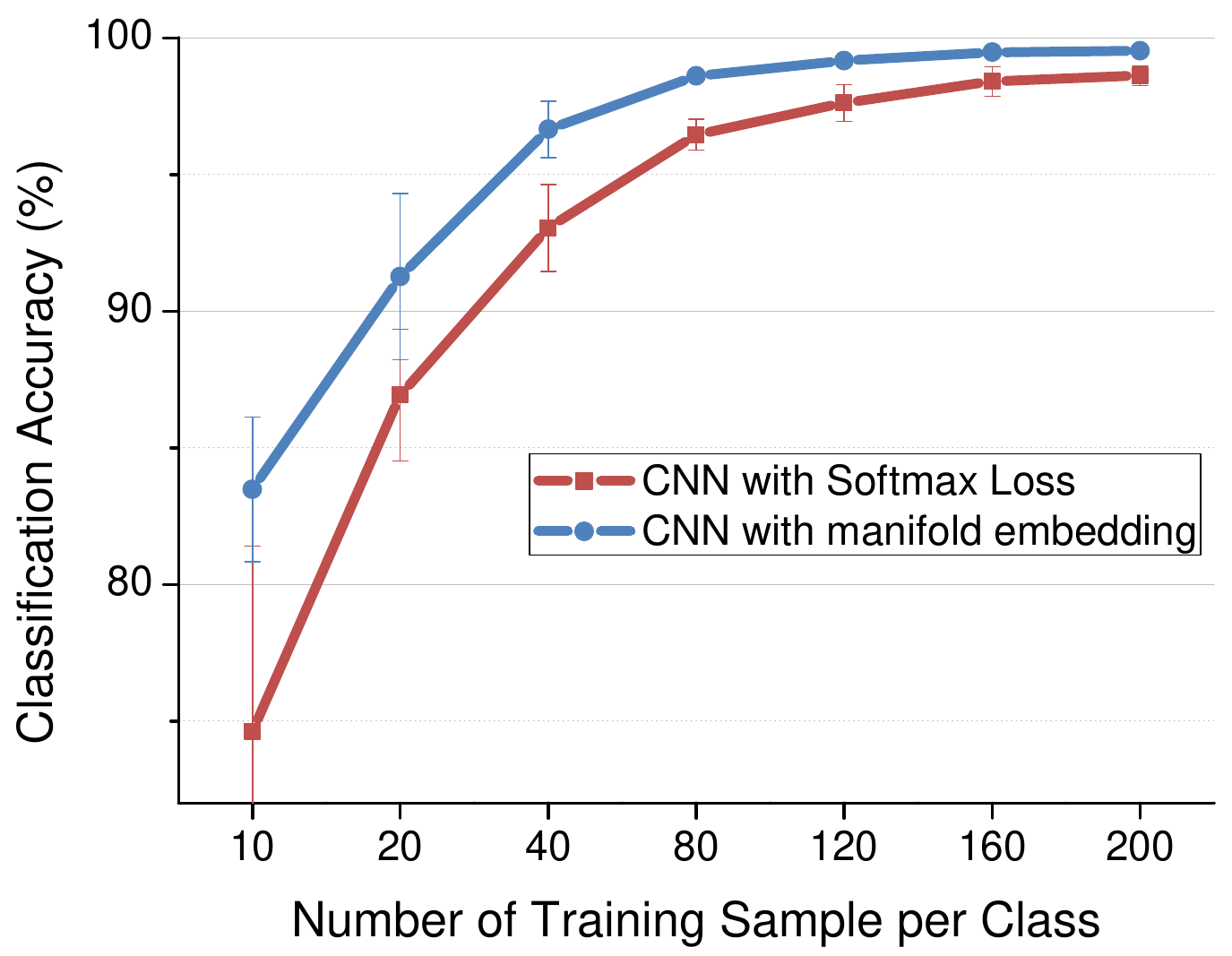}}
 \subfigure[]{\label{fig:number_salinas}\includegraphics[width=0.48\linewidth]{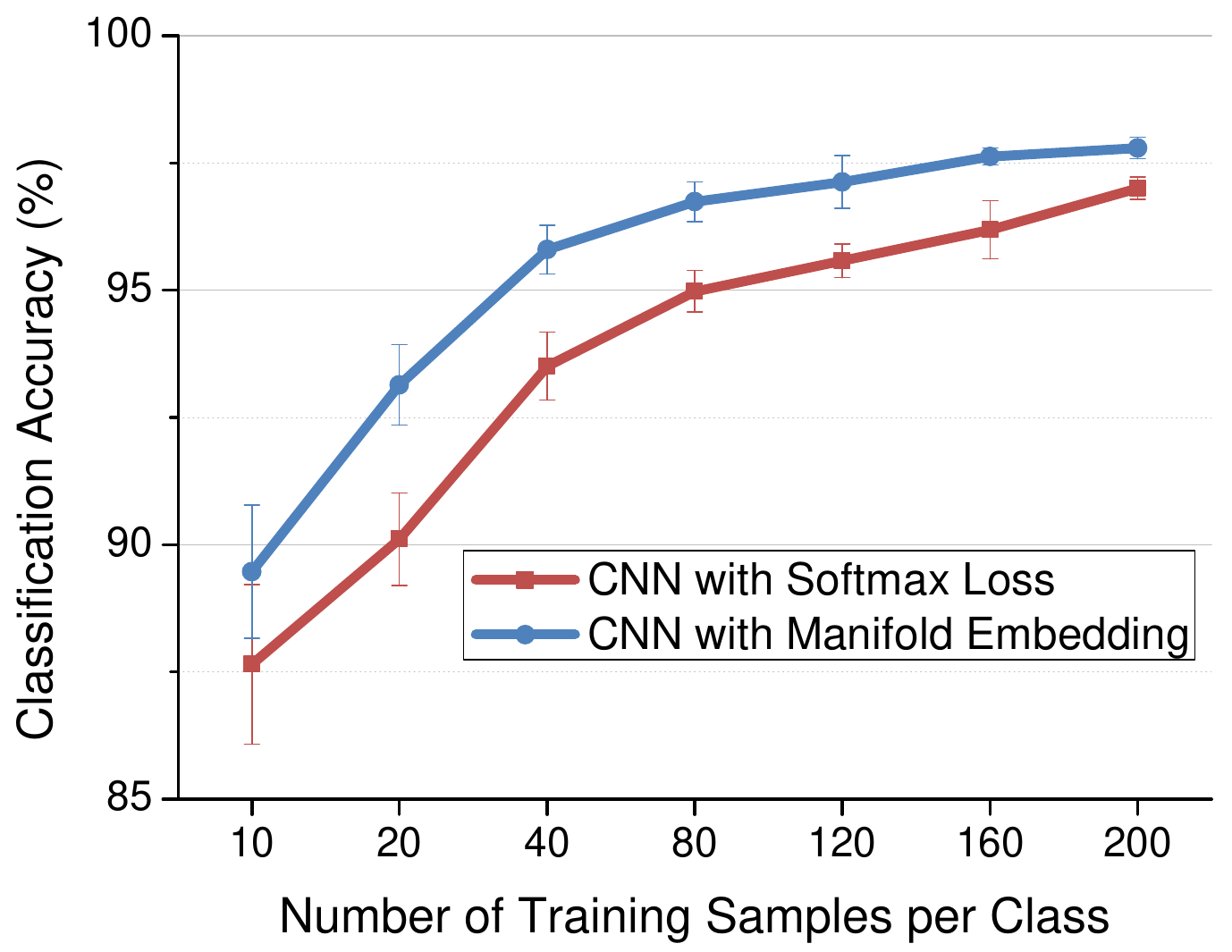}}
   \caption{{Classification performance of the proposed method under different number of training samples}  over {(a) Pavia University; (b) Salinas scene data.}}
\label{fig:number}
\end{figure}

\begin{figure}[t]
\centering
 \subfigure[]{\label{fig:mcnemar_pavia}\includegraphics[width=0.48\linewidth]{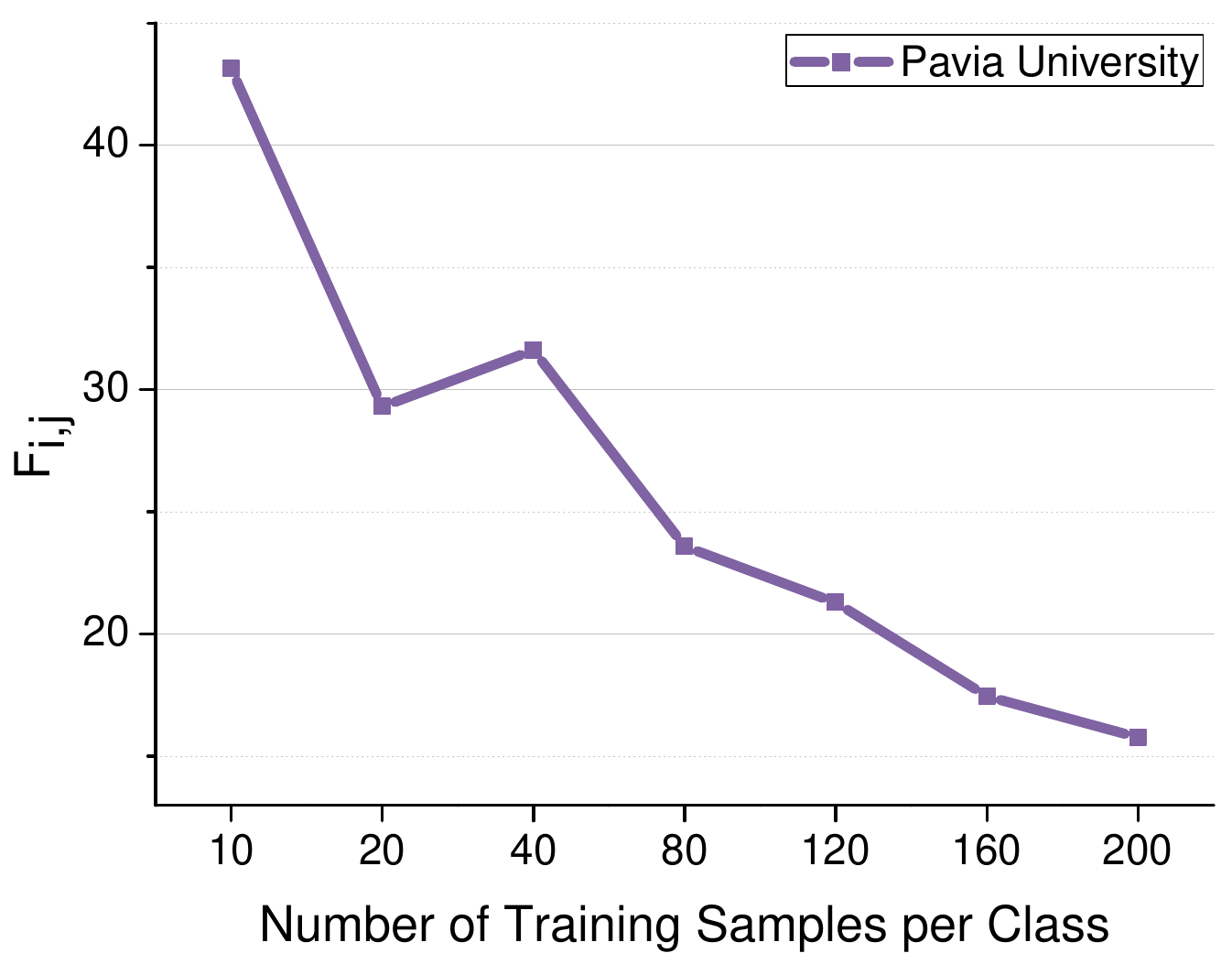}}
 \subfigure[]{\label{fig:mcnemar_salinas}\includegraphics[width=0.48\linewidth]{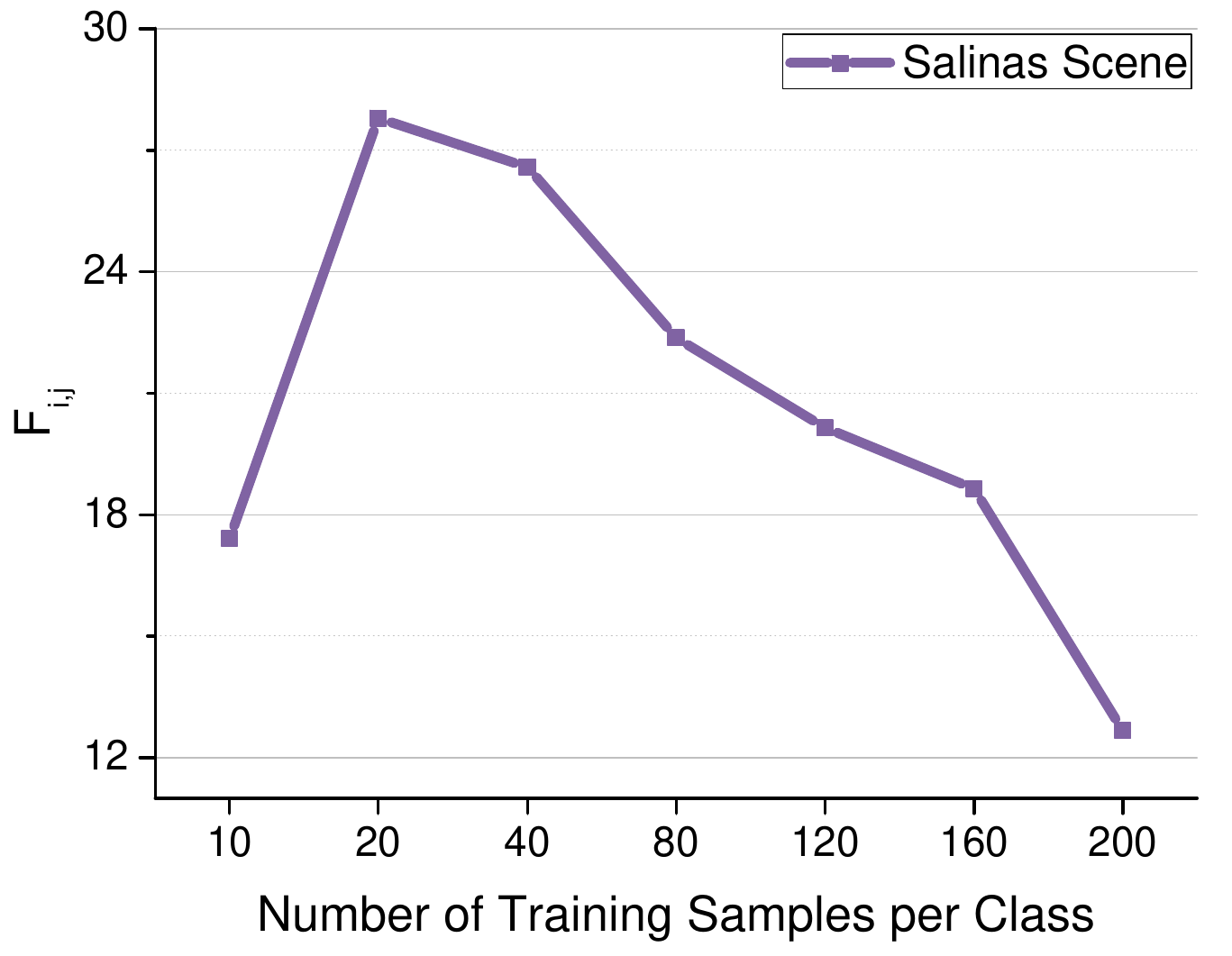}}
   \caption{{The value of Mcnemar's test between the CNN trained with deep manifold embedding and the softmax loss under different number of training samples over (a) Pavia University; (b) Salinas scene data.}}
\label{fig:mcnemar}
\end{figure}

Fig. \ref{fig:number} shows the classification performance of the developed method with different number of training samples and Fig. \ref{fig:mcnemar} shows the corresponding Mcnemar's test value $|F_{ij}|$ between the CNN trained with the proposed method and the CNN trained with the softmax loss only. From the figures, we can obtain the following conclusions.
\begin{enumerate}
\item The developed manifold embedding method can take advantage of the data manifold property in deep learning process and preserve the manifold structure in the low dimensional features which can improve the representational ability of the CNN model. Fig. \ref{fig:number} shows that the proposed method obtains better performance over all the three datasets under different number of training samples. Moreover, Fig. \ref{fig:mcnemar} also shows that the corresponding Mcnemar's test value over the three datasets is higher than 1.96 which means that the improvement of the proposed method is significant in statistic sense.
\item With the decrease of the training samples, the effectiveness of the developed method would be limited. Fig. \ref{fig:number} shows that the curves of the classification accuracy over each data set tend to be close to each other. Besides, from the Fig. \ref{fig:mcnemar}, it can be find that when the samples is limited, the value $|F_{ij}|$ is fluctuate which indicates that the effectiveness is negatively affected by the limited number of training samples. Just as the former subsection shows, this is because that constructing the data manifold requires a certain number of training samples. In contrary, too few samples may construct the false data manifold and show negative effects on the performance.
\end{enumerate}

\subsubsection{Effects of the Number of Sub-Classes $k$}
\begin{figure}[t]
\centering
 \subfigure[]{\label{fig:class_pavia}\includegraphics[width=0.48\linewidth]{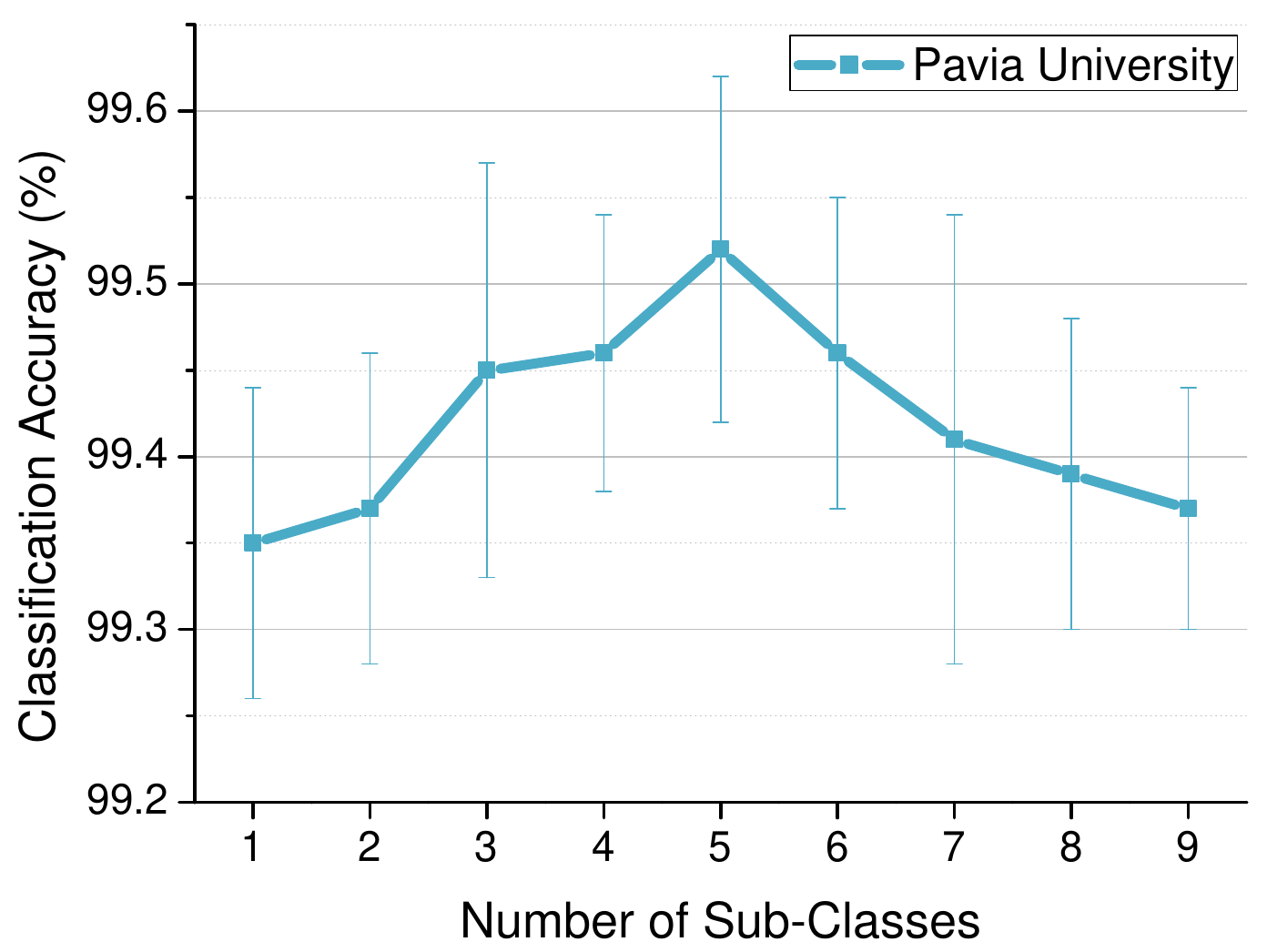}}
 \subfigure[]{\label{fig:class_salinas}\includegraphics[width=0.48\linewidth]{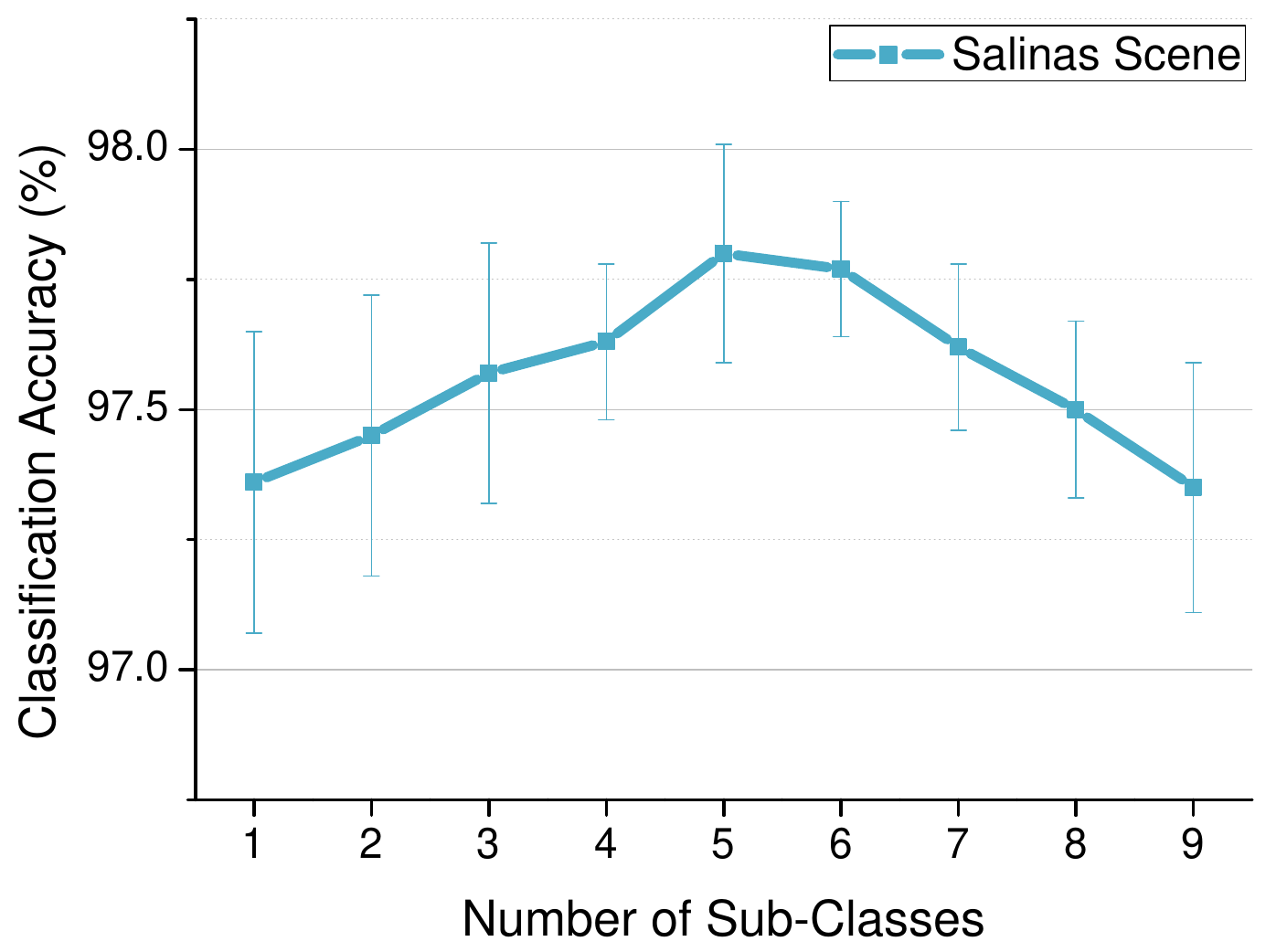}}
   \caption{Classification performance of the proposed method under different choices of the number of sub-classes  over (a) Pavia University; (b) Salinas scene data.}
\label{fig:classes}
\end{figure}

This part will show the performance of the developed method under different $k$.
{As showed in subsection \ref{subsec:manifold_modelling}, each class in the hyperspectral image is modelled as a certain nonlinear manifold and then divided into $k$ sub-classes. The number of the sub-classes demonstrates the nonlinearity of the manifold. When $k=1$, it means that all the samples in the class are linear and with the increase of the value of $k$, the nonlinearity of the manifold increases.}
In the experiments, the $k$ is chosen from $\{1,2,3,4,5,6,7,8,9\}$. The parameter $b$ is set to 5. Fig. \ref{fig:classes} presents the experimental results over the three data sets, respectively.

From the figure, we can find that a proper $k$ can guarantee a good performance of the developed manifold embedding method. From Fig. \ref{fig:class_pavia}, it can be find that when $k=5$, the classification accuracy over Pavia University data can achieve 99.52\% OA while $k=1$ can only lead to an accuracy of $99.35\%$ OA.  Besides, as Fig. \ref{fig:class_salinas}shows, for Salinas Scene data, when $k=5$, the proposed method performs the best. Generally, cross validation can be applied to select a proper $k$ in real-world application.

\subsubsection{Effects of the Number of neighbors $b$}

\begin{figure}[t]
\centering
 \subfigure[]{\label{fig:neighbor_pavia}\includegraphics[width=0.48\linewidth]{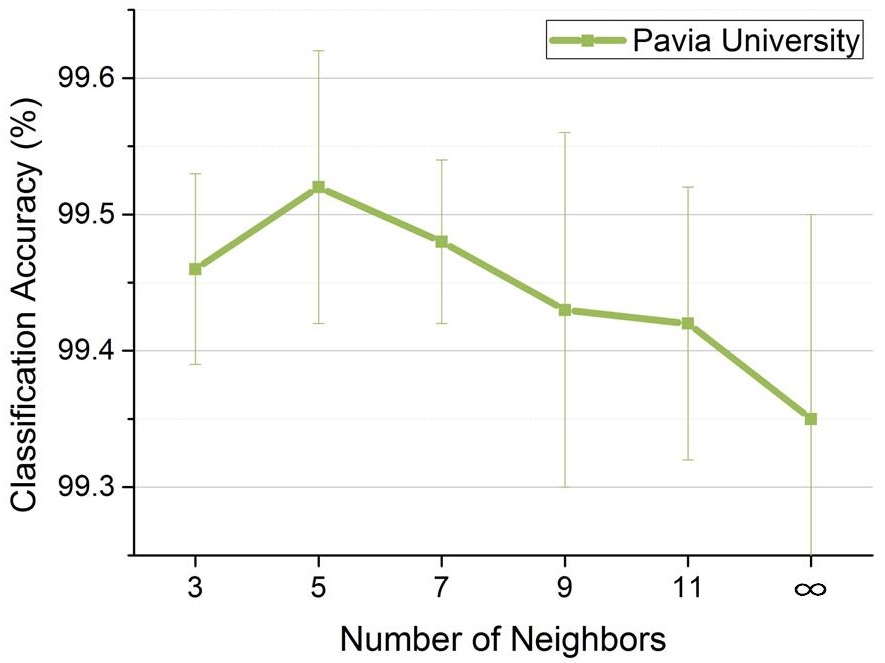}}
 \subfigure[]{\label{fig:neighbor_salinas}\includegraphics[width=0.48\linewidth]{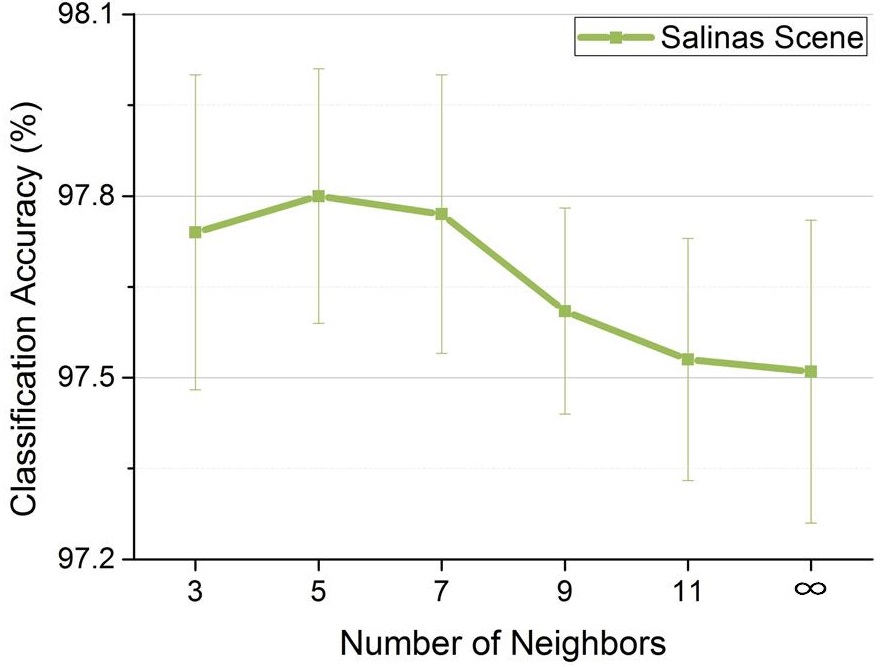}}
   \caption{{Classification performance of the proposed method under different choices of the number of neighbors  over (a) Pavia University; (b) Salinas scene data. ``$\infty$'' represents that all the samples are measured by Euclidean distance.}}
\label{fig:neighbors}
\end{figure}

Just as the parameter $k$, the number of neighbors $b$ also plays an important role in the developed method. Generally, extremely small $b$, such as $b=1$, would lead to the extremely ``steep'' of the constructed data manifold. While extremely large $b$ would lead to the overly smoothness of the data manifold. This subsection would discuss the performance of the developed method under different number of neighbors $b$. In the experiments, the $b$ is chosen from $\{3,5,7,9,11\}$. We also present the results when $b$ approaches infinity, namely all the samples are measured by Euclidean distance. In this set of experiments, the parameter $k$ is set to 5. Fig. \ref{fig:neighbors} shows the classification results of the proposed method under different choices of $b$ over the two data sets, respectively.
Inspect the tendencies in Fig. \ref{fig:neighbors} and we can note that the following hold.

Firstly, different $b$ can also significantly affect the performance of the developed method. Coincidentally, the performance of the proposed method achieves the best performance when $b$ is set to 5. Besides, the application of the Geodesic distance other than the Euclidean distance can improve the performance of the deep manifold embedding method. As Fig. \ref{fig:neighbor_pavia} shows, the proposed method can achieve 99.52\% over Pavia University data under Geodesic distance which is higher than 99.35\% under Euclidean distance. From Fig. \ref{fig:neighbor_salinas}, it can also be noted that over Salinas scene data, the proposed method under Geodesic distance can achieve 97.80\% which is better than 97.51\% under Euclidean distance.

\subsubsection{Comparisons with the Samples-based Loss}

This work also compares the developed deep manifold embedding with other recent {samples-based losses}. Here, we choose three representative {losses in prior works as baselines}, namely the softmax loss, center loss \cite{center}, and  structured loss \cite{lifted}. Table \ref{table:comparison_sample_class} lists the comparison results over the two data sets, respectively.

From the table, we can find that the proposed deep manifold embedding which can take advantage of the data manifold property within the hyperspectral image and preserve the {manifold structure of each class} in the low dimensional features can be more fit for the classification task than these samples-based loss. Over the Pavia University data, the proposed method can obtain an accuracy of 99.52\% outperform the CNN trained with the softmax loss (98.61\%), center loss (99.28\%), and the structured loss (99.27\%). Over the Salinas Scene data, the proposed method also outperforms these prior samples-based loss (see the table for details).

\begin{table}[t]
\begin{center}
\caption{Comparisons with other sample-wise loss. This work selects the softmax loss, the center loss \cite{center} and the structured loss \cite{lifted}. PU, SA stands for the Pavia University, and the Salinas Scene data, respectively.}
\label{table:comparison_sample_class}
\begin{tabular}{| c | c | c | c | c | c |}
\hline
{\bf Data}     &   {\bf Methods}     &  {\bf OA(\%)} &  {\bf AA(\%)} &  {\bf KAPPA(\%)} & $F_{ij}$ \\
\hline\hline
\multirow{4}{*}{\textbf{PU}}& {\bf Softmax Loss}    &  $98.61$ &  ${98.56}$  &  ${98.14}$  & 15.77 \\
& {\bf Center Loss}    &  $99.28$ &  $99.13$  &  $99.03$  & 6.03 \\
& {\bf Structured Loss} &  $99.27$ &  $99.12$  &  $99.02$  & 6.22 \\
& {\bf Proposed Method } &  $\mathbf{99.52}$ &  $\mathbf{99.37}$  &  $\mathbf{99.35}$  & $-$ \\
 \hline\hline
\multirow{4}{*}{\textbf{SA}} &{\bf Softmax Loss}    &  $97.01$ &  $98.73$  &  $96.65$  & 12.67 \\
&  {\bf Center Loss}    &  $97.43$ &  ${98.95}$  &  ${97.12}$  & 6.42 \\
&{\bf Structured Loss}    &  $97.40$ &  ${98.93}$  &  ${97.09}$  &  7.05 \\
&  {\bf Proposed Method} &  $\mathbf{97.80}$ &  $\mathbf{99.09}$  &  $\mathbf{97.54}$  & $-$ \\
\hline

\end{tabular}
\end{center}
\end{table}

\begin{figure*}[t]
\centering
 \subfigure[]{\label{fig:6a}\includegraphics[width=0.14\linewidth]{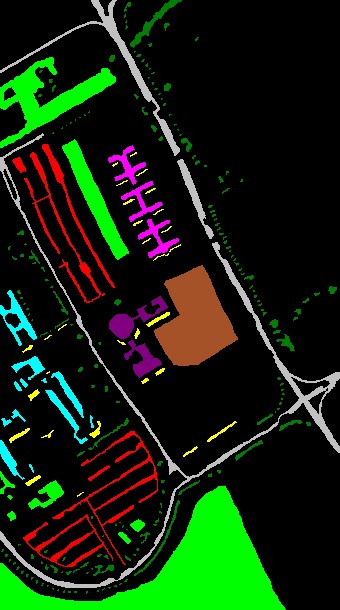}}
 \subfigure[]{\label{fig:6b}\includegraphics[width=0.14\linewidth]{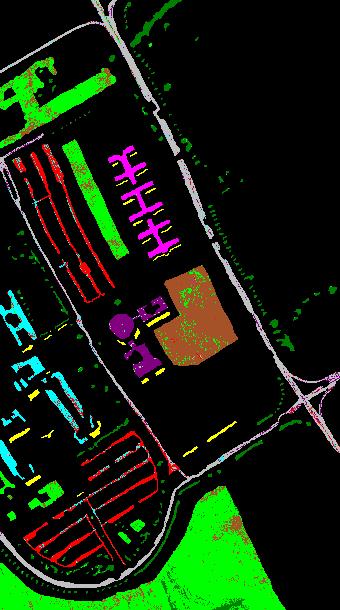}}
 \subfigure[]{\label{fig:6c}\includegraphics[width=0.14\linewidth]{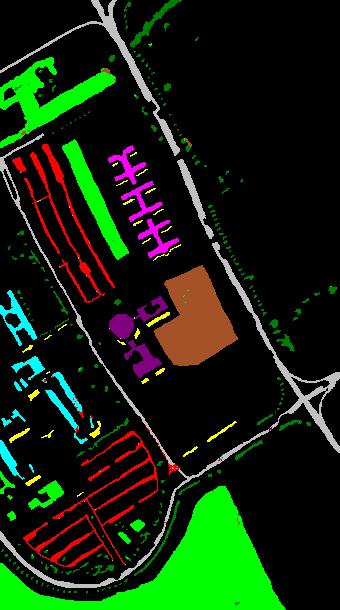}}
 \subfigure[]{\label{fig:6d}\includegraphics[width=0.14\linewidth]{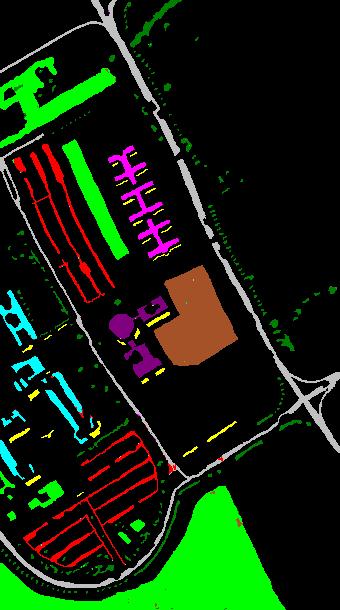}}
 \subfigure[]{\label{fig:6e}\includegraphics[width=0.14\linewidth]{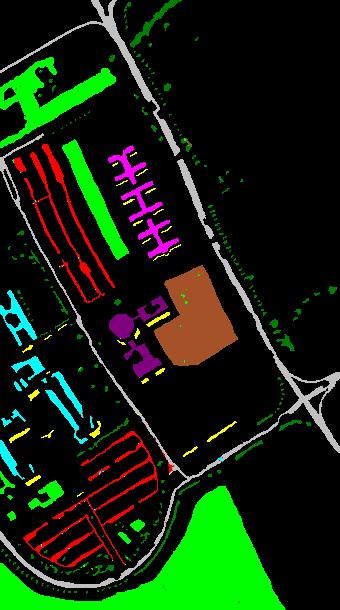}}
 \subfigure[]{\label{fig:6f}\includegraphics[width=0.14\linewidth]{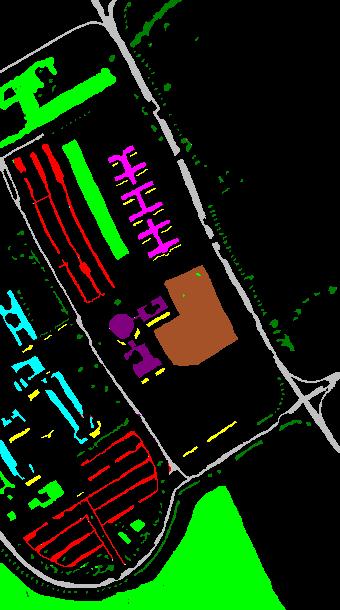}}
 \subfigure[]{\includegraphics[width=0.1\linewidth]{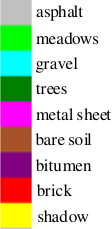}}
   \caption{Pavia University classification maps by different methods with 200 samples per class for training (overall accuracies). (a) groundtruth; (b) SVM (86.54\%); (c) CNN with softmax loss (98.91\%); (d) CNN with center loss (99.25\%) ; (e) CNN with structured loss (99.42\%); (f) CNN with developed manifold embedding loss (99.66\%); (g) map color.}
\label{fig:pavia_map}
\end{figure*}

\begin{figure*}[t]
\centering
 \subfigure[]{\label{fig:8a}\includegraphics[width=0.16\linewidth]{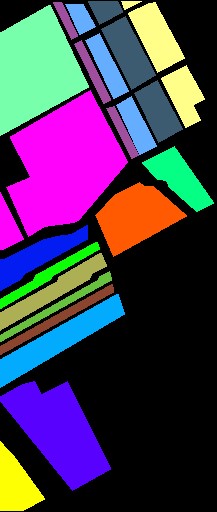}}
 \subfigure[]{\label{fig:8b}\includegraphics[width=0.16\linewidth]{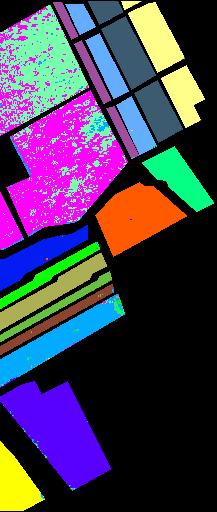}}
 \subfigure[]{\label{fig:8c}\includegraphics[width=0.16\linewidth]{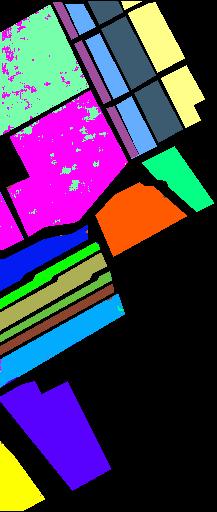}}
 \subfigure[]{\label{fig:8d}\includegraphics[width=0.16\linewidth]{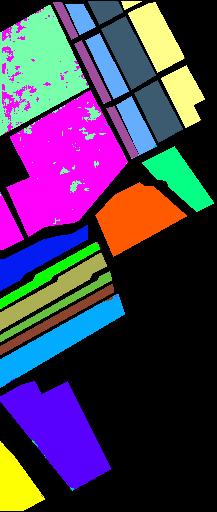}}
 \subfigure[]{\label{fig:8e}\includegraphics[width=0.16\linewidth]{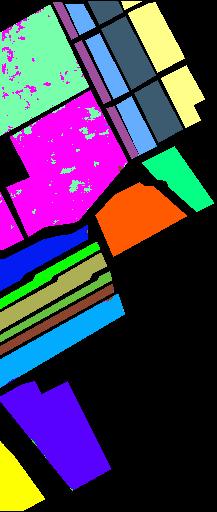}}
 \subfigure[]{\label{fig:8f}\includegraphics[width=0.16\linewidth]{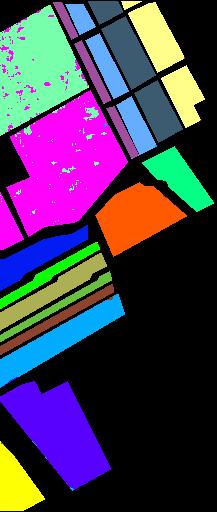}}
 \subfigure[]{\includegraphics[width=0.9\linewidth]{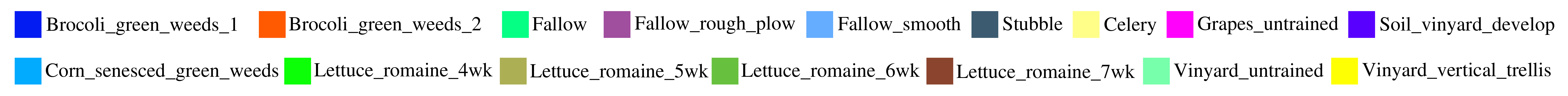}}
   \caption{Salinas Scenes classification maps by different methods with 200 samples per class for training (overall accuracies). (a) groundtruth;  (b) SVM (90.69\%); (c) CNN with softmax loss (97.06\%); (d) CNN with center loss (97.05\%); (e) CNN with structured loss (97.41\%); (f) CNN with developed {manifold embedding loss} (98.03\%); (g) map color.}
\label{fig:salinas_map}
\end{figure*}

Furthermore, we present the classification maps in Fig. \ref{fig:pavia_map}, and \ref{fig:salinas_map} by different methods over the Pavia University, and  Salinas Scene data, respectively. Compare Fig. \ref{fig:6c} and \ref{fig:6f}, \ref{fig:8c} and \ref{fig:8f}, and it can be easily noted that the CNN model trained with the deep manifold embedding can improve the performance of the CNN model. Besides, compare Fig. \ref{fig:6d} and \ref{fig:6f}, \ref{fig:8d} and \ref{fig:8f}, and we can find that the deep manifold embedding which can take advantage of the manifold structure can better model the hyperspectral image than the center loss. When compared \ref{fig:6e} and \ref{fig:6f}, \ref{fig:8e} and \ref{fig:8f}, we can also note that the proposed method can significantly decrease the classification errors obtained by the structured loss.

\subsubsection{Comparisons with the State-of-the-Art Methods}

{To further validate the effectiveness of the proposed manifold embedding method for hyperspectral image classification, we further make comparison with a number of the state-of-the-art methods: CNN-PPF \cite{15}, Contextual DCNN \cite{04}, Deep Spec-Spat \cite{19}, CDSAE \cite{06}, DPP-DML-MS-CNN \cite{gong_1}. Tables \ref{table:comparison_pavia}, and \ref{table:comparison_salinas} list the comparison results under the same experimental setups over the two data sets, respectively. It should be noted that the results in these tables are from the literatures where the method was first developed.}

Over Pavia University data, the developed method can obtain 99.52\% OA outperforms  CNN-PPF (96.48\% OA) \cite{15}, Contextual DCNN (97.31\% OA) \cite{04}, Deep Spec-Spat (95.58\% OA) \cite{19}, and DPP-DML-MS-CNN (99.46\% OA) \cite{gong_1}. Besides, over Salinas Scene data, the developed method can also provide competitive results (see tables \ref{table:comparison_salinas} for detail).
{Furthermore, the recently developed manifold-based methods are also compared with the proposed method for hyperspectal image classification and the proposed DMEM method can provide a comparable or even better performance than these methods (e.g. LMSDL \cite{duan2020}).}
{It can also be noted from the tables that the used deep model contains less or comparable number of parameters (about 2.5M) but better performance when compared with that of other deep models (e.g., about 18M parameters of deep spec-spat \cite{19}, 0.3M of CNN-PPF \cite{15}, 3.5M of DPP-DML-MS-CNN \cite{gong_1})}.

\begin{table*}[t]
\begin{center}
\caption{Classification performance of different methods over Pavia {University} data in the most recent literature ({200 samples per class} for training) where `param' means the number of parameters of the methods. }
\label{table:comparison_pavia}
\begin{tabular}{| c | c|c| c | c | c |}
\hline
{\bf Methods}    & {\bf Param.} & {\bf Training time (s)}   &  {\bf OA(\%)} &  {\bf AA(\%)} &  {\bf KAPPA(\%)} \\
\hline\hline

{\bf CNN-PPF \cite{15}}  & 0.26M & 14400 &  $96.48$ &  $97.03$  &  $-$\\ 
 {\bf Contextual DCNN \cite{04}} &0.60M& 751&  $97.31$ &  $95.77$  &  $-$\\   
  {\bf Deep Spec-Spat \cite{19}} & 18.72M& $-$  &  $95.58$ &  ${{98.41}}$  &  ${94.15}$\\
  {{\bf CDSAE \cite{06}}}   &0.026M& 6480&  {$97.59$} &  {${97.66}$}  & {${96.76}$}\\ 
{{\bf DPP-DML-MS-CNN \cite{gong_1}}}  & 3.38M& 680 &  ${{99.46}}$ &  ${\mathbf{99.39}}$  &  ${{99.27}}$\\  
  {\bf Proposed Method} & 2.68M & 2196& $\mathbf{99.52}$ &  ${99.37}$  &  $\mathbf{99.35}$\\   
\hline

\end{tabular}
\end{center}
\end{table*}

%
%

\begin{table*}[t]
\begin{center}
\caption{Classification performance of different methods over Salinas Scene data in the most recent literature({200 samples per class} for training) where `param' means the number of parameters of the methods. }
\label{table:comparison_salinas}
\begin{tabular}{| c | c | c | c | c | c |}
\hline
{\bf Methods}   & {\bf Param.} & {\bf Training time (s)} &  {\bf OA(\%)} &  {\bf AA(\%)} &  {\bf KAPPA(\%)} \\
\hline\hline

{\bf CNN-PPF \cite{15}}    & 0.29M & 43200 & $94.80$ &  $97.73$  &  $-$\\ 
 {\bf Contextual DCNN \cite{04}} & 1.83M & 896 &  $95.07$ &  $98.28$  &  $-$\\  
 {\bf Deep Spec-Spat \cite{19}}    & 19.84M & $-$ &  $97.58$ &  ${{97.33}}$  &  ${97.29}$\\
  {{\bf CDSAE \cite{06}}}    & 0.056M & 5400 & {$96.07$} &  {${97.56}$}  & {${96.78}$}\\ 
{{\bf DPP-DML-MS-CNN \cite{gong_1}}} & 3.65M & 642 &  ${97.51}$ &  ${98.85}$  &  ${\mathbf{97.88}}$\\  
  {\bf Proposed Method} & 2.73M & 2965&  $\mathbf{97.80}$ &  $\mathbf{99.09}$  &  ${97.54}$\\  
\hline

\end{tabular}
\end{center}
\end{table*}

{In addition, the proposed method also provide an impressive performance under limited number of training samples.  For example, over Pavia University data, the proposed method can obtain an accuracy of 96.65\% under 40 training samples per class outperforms the ACR \cite{yu2020} (less than 90\%). 

To sum up, the joint supervision of the developed manifold embedding loss and softmax loss can always enhance the deep models' ability to extract discriminative representations and obtain comparable or even better results when compared other state-of-the-art methods.}
\subsection{{Results over Houston 2013 and 2018 Dataset}}

{The results over Pavia University, and Salinas Scene data have thoroughly analyzed the performance of the proposed method.
In this set of experiments, we will further conduct additional experimental results on more challenging datasets, namely Houston 2013 and 2018 Dataset, to show the effectiveness of the proposed method.

The experimental results of Houston 2018 data have been shown in table \ref{table:houston2018}. From the table, we can find that the proposed method which can obtain an accuracy of $82.62\%$ outperforms vanilla CNNs (78.47\%). This means that the proposed DMEM method can significantly improve the performance of the learned deep model.

\begin{table}[t]
\begin{center}
\caption{{Classification accuracies ($Mean\pm SD$) (OA, AA, and Kappa) of different methods achieved on the Houston 2018 data. } }
\label{table:houston2018}
\begin{tabular}{|c | c | c c c|}
\hline
\multicolumn{2}{|c|}{\bf Methods}     &  {\bf SVM-POLY} &  {\bf CNN} &  {\bf Proposed Method} \\
\hline\hline
\multirow{20}{*}{\rotatebox{90}{\tabincell{c}{\textbf{Classification} \\ \textbf{Accuracies (\%)}}}}          & C1   &  $34.92$  &  ${{60.20\pm 26.5}}$ &  $\mathbf{66.39\pm 5.10}$\\
                                                                                                             & C2   &  $57.27$  &  ${85.86\pm 3.11}$ &  $\mathbf{87.64\pm 2.07}$\\
                                                                                                             & C3   &  $\mathbf{100.0}$  &  $99.71\pm 0.00$ &  ${96.74\pm 0.60}$\\
                                                                                                             & C4   &  $81.40$  &  $\mathbf{93.94\pm 0.47}$ &  ${91.44\pm 1.32}$\\
                                                                                                             & C5   &  $25.18$  &  $\mathbf {62.48\pm 11.3}$ &  ${47.29\pm 5.13}$\\
                                                                                                             & C6   &  $28.85$  &  $\mathbf{32.06\pm 5.77}$ &  ${28.51\pm 1.46}$\\
                                                                                                             & C7   &  $0$  &  $\mathbf{58.82\pm 40.4}$ &  ${30.25\pm 0.00}$\\
                                                                                                             & C8   &  $\mathbf{82.76}$  &  ${81.00\pm 1.47}$ &  ${76.83\pm 9.76}$\\
                                                                                                             & C9   &  $54.61$  &  $45.60\pm 1.11$ &  $\mathbf{60.28\pm 1.82}$\\
                                                                                                             & C10   &  $41.81$  &  $\mathbf{46.24\pm 8.39}$ &  ${42.12\pm 6.31}$\\
                                                                                                             & C11   &  $\mathbf{53.86}$  &  ${41.04\pm 14.3}$ &  ${39.10\pm 5.24}$\\
                                                                                                             & C12   &  $1.24$  &  $\mathbf{7.17\pm 7.07}$ &  ${0.00\pm 0.00}$\\
                                                                                                             & C13   &  $41.61$  &  ${55.62\pm 5.22}$ &  $\mathbf{56.84\pm 3.11}$\\
                                                                                                             & C14   &  $17.41$  &  $\mathbf{45.41\pm 5.35}$ &  ${39.72\pm 4.45}$\\
                                                                                                             & C15   &  $19.99$  &  $\mathbf{56.10\pm 2.28}$ &  ${53.43\pm 1.89}$\\
                                                                                                             & C16   &  $28.22$  &  $\mathbf{79.94\pm 0.05}$ &  ${73.05\pm 2.73}$\\
                                                                                                             & C17   &  $\mathbf{100.0}$  &  ${93.18\pm 9.64}$ &  ${90.91\pm 6.43}$\\
                                                                                                             & C18   &  $21.31$  &  $84.67\pm 7.02$ &  $\mathbf{50.03\pm 36.11}$\\
                                                                                                             & C19   &  $41.40$  &  $\mathbf{85.53\pm 5.52}$ &  ${79.05\pm 8.87}$\\
                                                                                                             & C20   &  $44.02$  &  $\mathbf{71.32\pm 15.9}$ &  ${59.90\pm 1.95}$\\
 \hline
 \multicolumn{2}{|c|}{{\bf OA}  (\%)}      &  $51.95$ &  $54.89\pm 1.36$  &  $\mathbf{60.12\pm 0.49}$\\
 \hline
 \multicolumn{2}{|c|}{{\bf AA}  (\%)}      &  $43.79$ &  $\mathbf{64.29\pm 2.20}$  &  ${58.33\pm 1.78}$\\
 \hline
 \multicolumn{2}{|c|}{{\bf KAPPA} (\%)}    &  $41.38$ &  $46.96\pm 1.34$  &  $\mathbf{51.00\pm 0.30}$\\
\hline
\end{tabular}
\end{center}
\end{table}

Furthermore, {table \ref{table:houston2013} presents the classification accuracies of the proposed method compared with other most recent methods over Houston 2013 data.} It can be noted from the table that the proposed method can obtain an overall accuracy of $86.98\%$ outperform the recently developed methods, such as Gabor-CNN (84.32\%) \cite{gabor}, SaCRT (83.69\%) \cite{structureaware}, and FreeNet (86.61\%) \cite{fpga}.
{Since the proposed method can also be seen as dimensionality reduction method, we compare the PCA+CNN where dimensionality reduction method PCA is using before CNN with the proposed method. Over the houston2013 data, the proposed method can obtain  an accuracy of 86.98\% OA which is higher than 84.82\% obtained by the PCA+CNN. Therefore, compared with other dimensionality reduction method, the proposed method can also provide a superior performance. }

\begin{table*}[t]
\begin{center}
\caption{{Classification accuracies ($Mean\pm SD$) (OA, AA, and Kappa) of SVM, 3D-CNN \cite{3dcnn}, SSRN \cite{zhongzl2017}, Gabor-CNN \cite{gabor}, CNN-PPF \cite{15}, SaCRT \cite{structureaware}, CapsNet \cite{capsulenet}, DpresNet \cite{dpresnet},  FreeNet \cite{fpga}, PCA + CNN and the proposed method achieved on the Houston 2013 data. }}
\label{table:houston2013}
\begin{tabular}{|c | c | c c c c c c c c c c c|}
\hline
\multicolumn{2}{|c|}{\bf Methods}     &  {\bf SVM} & {\bf 3D-CNN} & {\bf Gabor-CNN}& {\bf CNN-PPF} & {\bf SSRN} & {\bf SaCRT}  & {\bf CapsNet} & {\bf DpresNet} &  {\bf FreeNet} & {\bf PCA+CNN} &  {\bf Proposed Method} \\
\hline\hline
\multirow{15}{*}{\rotatebox{90}{\tabincell{c}{\textbf{Classification} \\ \textbf{Accuracies (\%)}}}}          & C1   &  $82.05$ & 80.24 & 82.30 & 83.00 & 82.62  & 83.00 & 82.09 & 80.59 & ${80.91}$ & 81.76 &  $\mathbf{83.10}$\\
                                                                                                             & C2   &  $80.55$ & 74.81 &  84.24 & 84.21 & ${85.90}$ & 82.80 & 85.15 & 82.67& ${84.21}$ & $\mathbf{96.62}$&  ${84.27}$\\
                                                                                                             & C3   &  $100.0$ & 58.01 & 95.61 & 100.0 & 98.02 & 100.0 & 94.65& 79.52 & $98.02$ & 99.60 &  $\mathbf{100.0}$\\
                                                                                                             & C4   &  $92.52$ & 91.75 & 92.39 & 92.61 &91.95  & 97.82 & 92.89& 86.08 & ${91.95}$ & $\mathbf{99.62}$ &  ${93.01}$\\
                                                                                                             & C5   &  $98.11$ & 92.61 & 99.80 & 99.81 & 100.0 & 98.58 & 99.91& 97.23 & $\mathbf {100.0}$ & 100.0 &  ${99.98}$\\
                                                                                                             & C6   &  $95.10$ & 93.00 & 96.47 & 95.10 & 95.80 & $\mathbf{99.30}$ & 94.16 &93.14 & ${96.50}$ & 95.80 &  ${97.06}$\\
                                                                                                             & C7   &  $75.00$ & 85.64 & 84.58& 83.77 & 84.42  & 87.59 & 83.30 & 80.97& $\mathbf{88.53}$ & 78.54 &  ${81.36}$\\
                                                                                                             & C8   &  $40.17$ & 55.56 & 73.09 & 46.53 & 58.50 & 62.39 & 62.77 & 70.06 & $\mathbf{74.83}$ & 63.24 &  ${74.32}$\\
                                                                                                             & C9   &  $74.88$  & 84.71 & 78.14 & 78.56& 91.88 & 82.91 & 79.61 &82.01 & $\mathbf{87.72}$ & 81.21 &  ${84.89}$\\
                                                                                                             & C10   &  $51.64$ & 43.14 & 58.01 &53.28 & 57.72 & 54.54 & 65.91 &52.45 & ${62.25}$ & 58.69 &  $\mathbf{69.44}$\\
                                                                                                             & C11   &  $78.37$ & 76.03 & 73.35 & 82.16& 71.54 & 78.37 & 86.05 &74.93 & $83.40$ & 86.81 &  $\mathbf{89.94}$\\
                                                                                                             & C12   &  $68.40$ & 81.75 & 86.74 &82.61 & 95.29 & 88.95 & 93.76 &93.57 & $\mathbf{98.84}$ & 83.77 & ${94.35}$\\
                                                                                                             & C13   &  $69.47$ & $\mathbf{90.52}$ & 91.16 & 80.35 & 87.72 & 80.70 & 88.77 &83.69 & ${88.42}$ & 79.30 &  ${81.68}$\\
                                                                                                             & C14   &  $100.0$ & 79.76 & 100.0 & 100.0 & 100.0 & 99.60 & 97.57 &89.27 & ${96.76}$ & 100.0 &  $\mathbf{100.0}$ \\
                                                                                                             & C15   &  $98.10$ & 89.47 & 77.73 & 99.58 & 99.79 & 98.31 & 96.19 & 96.03 & $94.29$ & 100.0 & $\mathbf{100.0}$\\
 \hline
 \multicolumn{2}{|c|}{{\bf OA}  (\%)}      &  $76.88$  & 79.73  & 84.32 & 84.11 & 84.04 & 83.69 & 85.96 & 81.62& $86.61$  & 84.82 & $\mathbf{86.98}$\\
 \hline
 \multicolumn{2}{|c|}{{\bf AA}  (\%)}      &  $80.29$ & 78.47  & 84.17  & 81.07 & 86.74 & 86.32 & 87.05 & 82.45 &${88.44}$  & 87.00 &  $\mathbf{88.89}$\\
 \hline
 \multicolumn{2}{|c|}{{\bf KAPPA} (\%)}    &  $75.13$ & 77.62  & 81.14 & 79.62 &  82.68 & 82.31 & 83.78 & 80.39 &$85.55$  & 83.56 & $\mathbf{85.93}$\\
\hline
\end{tabular}
\end{center}
\end{table*}

\begin{figure}[t]
\centering
\includegraphics[width=0.85\linewidth]{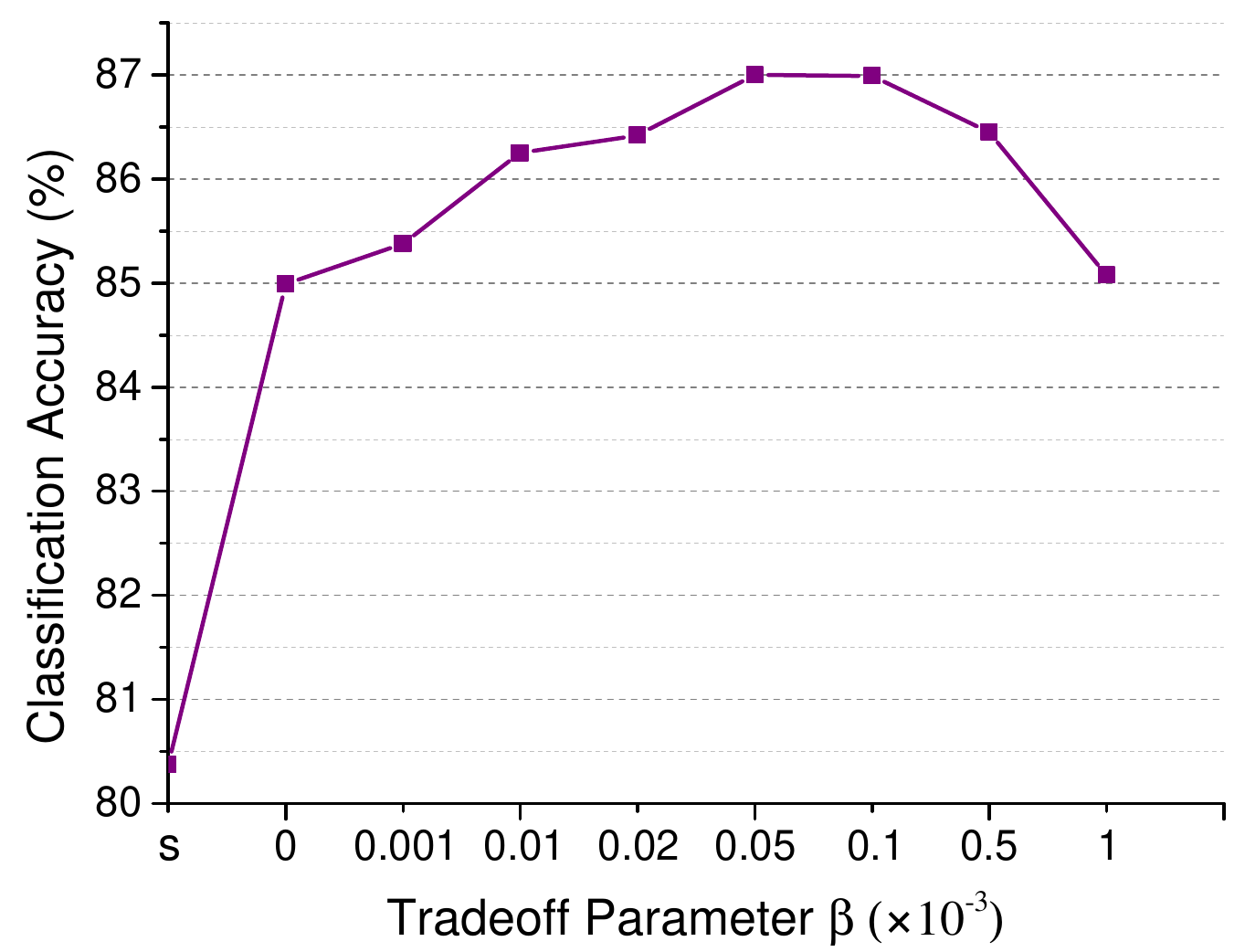}
   \caption{{Classification performance with different diversity tradeoff parameter $\beta$ over Houston 2013 data. 's' means the classification performance without DMEM loss}}
\label{fig:diversity_tradeoff}
\end{figure}

For the proposed method, the diversity tradeoff parameter $\beta$ also plays an important role in the training performance. Fig. \ref{fig:diversity_tradeoff} has shown the classification performance with different diversity tradeoff parameter $\beta$ over Houston 2013 data. It can be noted that the accuracy can achieve $86.99\%$ when $\beta$ is set to 0.0001 and $87\%$ when $\beta$ is set to 0.0005 which ranks the best. Based on Fig. \ref{fig:diversity_tradeoff}, we can find that when the diversity tradeoff parameter $\beta$ is set to 0, the trained model can obtain an accuracy of 84.99\% and the performance of the proposed method with the help of the diversity method. While excessively large $\beta$ can make the unstable of the training process (i.e., $\beta = 0.001$). Other datasets show the similar tendencies and therefore in the experiments, the diversity tradeoff parameter $\beta$ is set to 0.0001.
{Furthermore, different values of balance parameter $\lambda$ are also tested to comprehensively present the performance of the proposed method over Houston 2013 data. The value of $\lambda$ is set to 0, 0.00001, 0.00005, 0.0001, 0.0005, 0.001, respectively and the accuracy of the learned model can achieve 82.9\%, 84.4\%, 85.3\%, 87.0\%, 86.4\%, 84.9\%, seperately. It can be noted that the larger lambda guarantees a better performance while excessively large $\lambda$ would decrease the performance, sometimes even make the model fail to converge.}

Besides, to show the role of the DMEM loss played in the training process, Fig. \ref{fig:loss_houston2013} has presented the tendencies of the training loss over Houston2013. Fig. \ref{fig:loss} shows the loss tendencies over iterations and Fig. \ref{fig:loss_rate} presents the proportion of DMEM loss to the overall loss. It can be noted that with the training of the deep model the DMEM loss shows heavier weight and it can help the training process to skip the local optimum and obtain a model with better performance. Therefore, the DMEM loss which utilizes the intrinsic manifold structure of the hyperspectral image can improve the training process and present better performance.

\begin{figure}[t]
\centering
 \subfigure[]{\label{fig:loss}\includegraphics[width=0.48\linewidth]{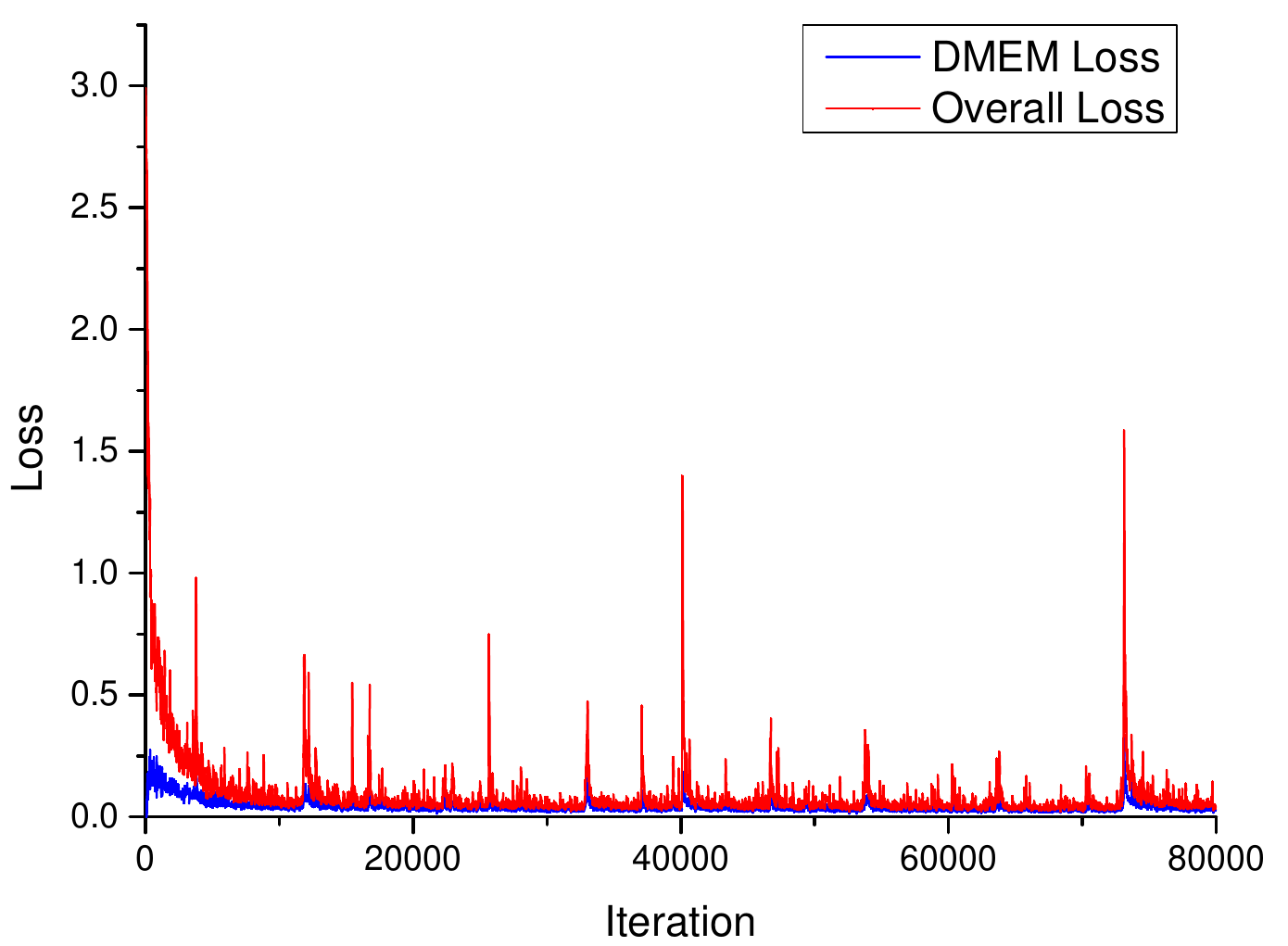}}
 \subfigure[]{\label{fig:loss_rate}\includegraphics[width=0.48\linewidth]{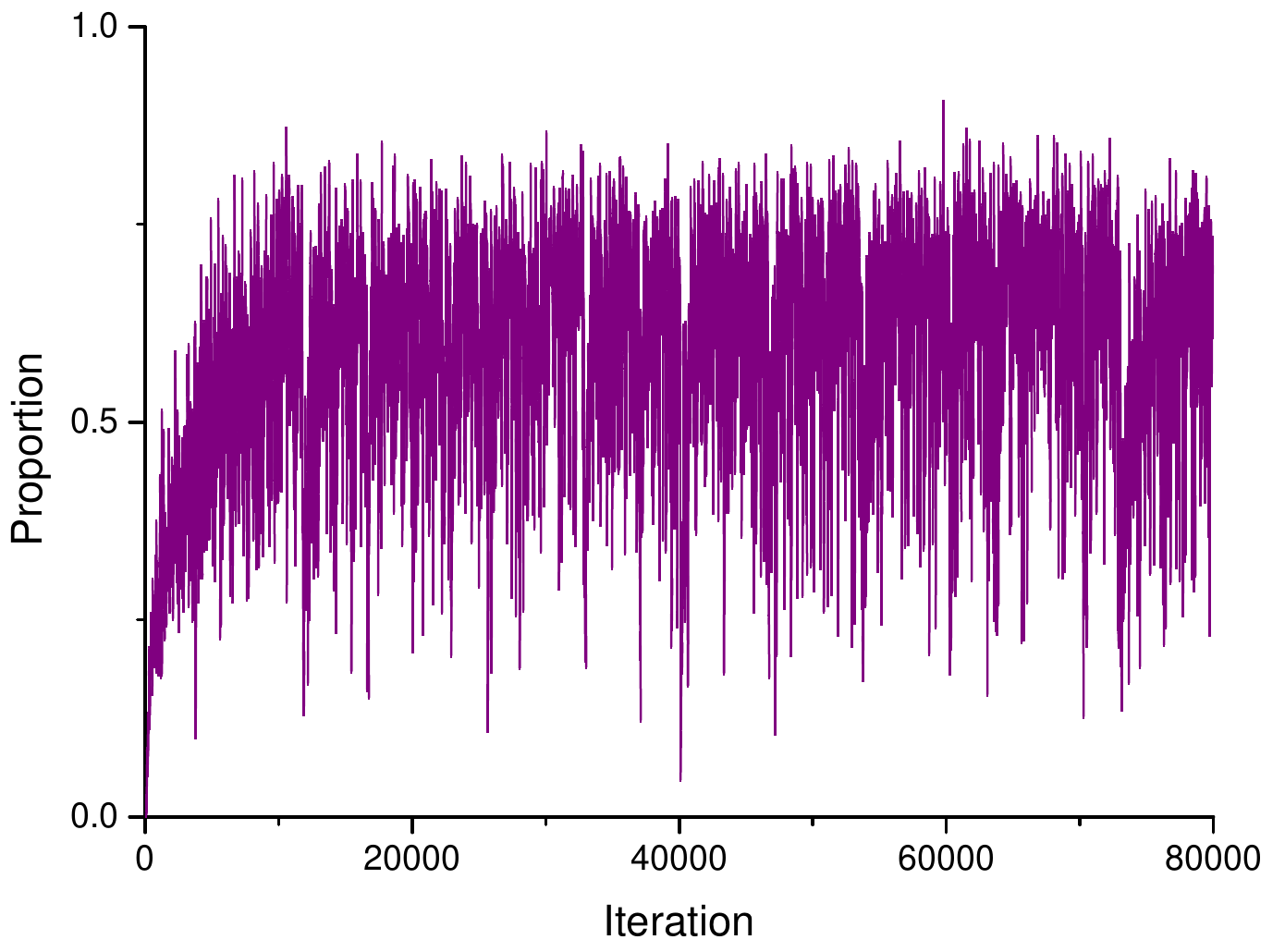}}
   \caption{{The tendencies of training loss over Houston2013 data. (a) Loss tendencies over iterations; (b) Proportion of DMEM loss to the overall loss.}}
\label{fig:loss_houston2013}
\end{figure}

We also test the performance of the proposed method with different spatial size over Houston 2013 data. The results have been shown in Fig. \ref{fig:spatial_size}. It can be find that when the spatial size is set to $7 \times 7$, the accuracy of the proposed method can be 87.63\%, which achieves the best. The purpose of the proposed method is to develop a novel deep learning method which can use the special intrinsic characteristics of the hyperspectral image to improve the training performance and therefore we select $5\times 5$ spatial size for experiments.
{In addition to the above-mentioned comparisons, we conduct additional experiments under different tradeoff parameter $\lambda$ over Houston 2013 data.}

\begin{figure}[t]
\centering
\includegraphics[width=0.85\linewidth]{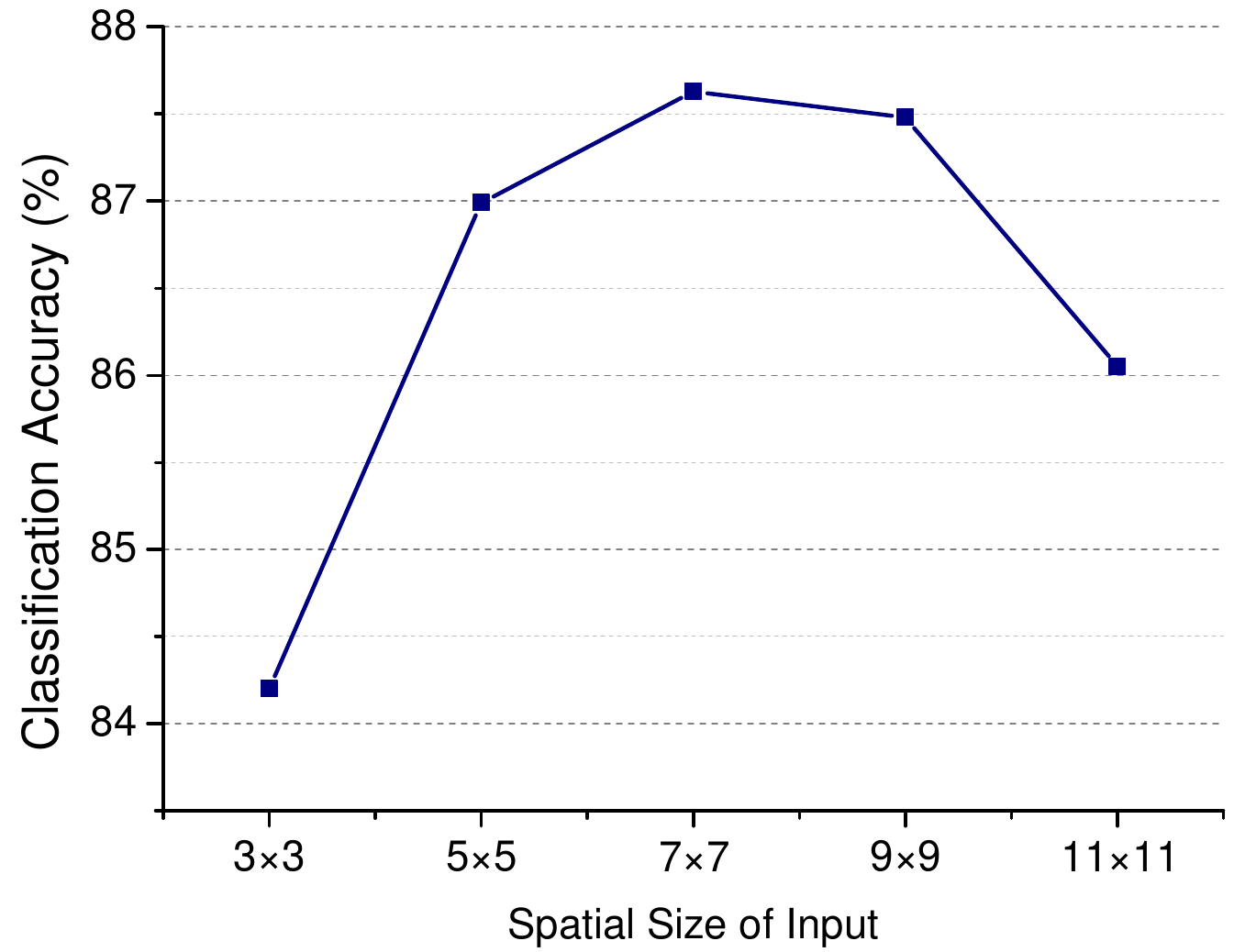}
   \caption{{Classification performance with different spatial size of input over Houston 2013 data.}}
\label{fig:spatial_size}
\end{figure}

Overall, the experiments on the Pavia University, and the Salinas Scene data as well as the more challenging data, such as the Houston 2013 and 2018, have shown the effectiveness of the proposed DMEM method to improve the effectiveness for deep learning in hyperspectral image classification.

}

\section{Conclusion and Discussion}\label{sec:conclusion}

Based on the intrinsic manifold structure of the hyperspectral image, this work develops a novel deep manifold embedding method (DMEM) for deep learning. This method aims to solve the
problem of general samples-based losses ignoring the intrinsic structure {of each class in} hyperspectral image.
Considering the characteristics of hyperspectral image, especially {the spectral signature}, the DMEM constructs {the training loss} using the manifold structure of the image and incorporates the {class-wise information} in the learning process. Therefore, the learned model by DMEM can be better fit for deep representation of the hyperspectral image.
The proposed method have been evaluated over {four
datasets} and compared to several related works to verify its performance and significance.
Using the intrinsic data structure in the deep learning process does help to improve the performance of the deep model and experimental results have validated the effectiveness of the developed DMEM.

As future work, it would be interesting to investigate the effectiveness of the manifold embedding on other {hyperspectral image processing tasks}, such as hyperspectral target detection. Besides, further consideration could be given to embed the manifold structure in other forms. Finally, {using} other data structures which can significantly affect the deep learning performance is another important future topic.


%



\section*{Acknowledgment}
The authors would like to thank the researchers for providing the Pavia University and the Salinas Scene data, and the National Center for Airborne Laser Mapping and the Hyperspectral Image Analysis Laboratory at the University of Houston for acquiring and providing the data, and the IEEE GRSS Image Analysis and Data Fusion Technical Committee.

\ifCLASSOPTIONcaptionsoff
  \newpage
\fi



%

\bibliographystyle{IEEEtran}
\bibliography{mybibfile}




%



\begin{IEEEbiographynophoto}{Zhiqiang Gong}
received the B.S. degree
in applied mathematics from Shanghai Jiao Tong
University, Shanghai, China, in 2013, the M.S.
degree in applied mathematics from the National
University of Defense Technology (NUDT),
Changsha, China, in 2015, and the Ph.D. degree
in information and communication engineering
from the National Key Laboratory of Science and
Technology on ATR, NUDT, in 2019.

He is currently an Assistant Professor with
the National Innovation Institute of Defense
Technology, Chinese Academy of Military Science, Beijing, China. He has
authored more than ten peer-reviewed articles in international journals, such as the
IEEE TRANSACTIONS ON NEURAL NETWORKS AND LEARNING SYSTEMS, the IEEE TRANSACTIONS ON CYBERNETICS,
the IEEE TRANSACTIONS ON GEOSCIENCE AND REMOTE SENSING,
the IEEE GEOSCIENCE AND REMOTE SENSING LETTERS, and the IEEE
JOURNAL OF SELECTED TOPICS IN APPLIED EARTH OBSERVATIONS AND
REMOTE SENSING. His research interests are computer vision, machine
learning, and image analysis.
\end{IEEEbiographynophoto}

\begin{IEEEbiographynophoto}{Weidong Hu}
received the B.S. degree in microwave
technology and the M.S. and Ph.D. degrees in communication
and electronic system from the National
University of Defense Technology (NUDT), Changsha,
China, in 1990, 1994, and 1997, respectively.

He is currently a Full Professor with the National
Key Laboratory of Science and Technology on ATR,
NUDT. His research interests include radar signal
and data processing.
\end{IEEEbiographynophoto}

\begin{IEEEbiographynophoto}{Xiaoyong Du}
was born in November 1976. He received the B.S. degree in applied mathematics, the M.S. degree in statistics, and the Ph.D. degree in information and communication engineering from the National University of Defense Technology, Changsha, China, in 1998, 2001, and 2005, respectively.

He is currently an Associate Professor with the ATR Laboratory, National University of Defense Technology, Changsha. His research interests include radar imaging and sparse signal processing.
\end{IEEEbiographynophoto}

\begin{IEEEbiographynophoto}{Ping Zhong (Senior Member, IEEE)}
received the M.S. degree in applied mathematics
and the Ph.D. degree in information and communication engineering from
the National University of Defense Technology (NUDT), Changsha, China,
in 2003 and 2008, respectively.

From March 2015 to February 2016, he is a Visiting Scholar in the
Department of Applied Mathematics and Theory Physics, University of
Cambridge, UK. He is currently a Professor with the National Key Laboratory
of Science and Technology on ATR, NUDT. He has authored more than 30
peer-reviewed papers in international journals such as the IEEE TRANSACTIONS
ON NEURAL NETWORKS AND LEARNING SYSTEMS, the IEEE
TRANSACTIONS ON IMAGE PROCESSING, the IEEE TRANSACTIONS
ON GEOSCIENCE AND REMOTE SENSING, and the IEEE JOURNAL OF
SELECTED TOPICS IN SIGNAL PROCESSING, and the IEEE JOURNAL
OF SELECTED TOPICS IN APPLIED EARTH OBSERVATIONS AND
REMOTE SENSING. His research interests include computer vision, machine
learning, and pattern recognition.

He was the recipient of the National Excellent Doctoral Dissertation Award
of China (2011) and New Century Excellent Talents in University of China
(2013). Dr. Zhong is a Referee of the IEEE TRANSACTIONS ON NEURAL
METWORKS AND LEARNING SYSTEMS, the IEEE TRANSACTION ON
IMAGE PROCESSING, IEEE TRANSACTIONS ON GEOSCIENCE AND
REMOTE SENSING, the IEEE JOURNAL OF SELECTED TOPICS IN
APPLIED EARTH OBSERVATIONS AND REMOTE SENSING, the IEEE
JOURNAL OF SELECTED TOPICS IN SIGNAL PROCESSING, and the
IEEE GEOSCIENCE AND REMOTE SENSING LETTERS.
\end{IEEEbiographynophoto}
%

%

%
\begin{IEEEbiographynophoto}{Panhe Hu}
was born in Shandong, China, in 1991. He received the B.S. degree from Xidian University, Xi’an, China, in 2013, and the Ph.D. degree from National University of Defense Technology (NUDT), Changsha, China, in 2019. He is currently a Lecturer with the College of Electronic Science and Technology, NUDT. His current research interests include radar system design, array signal processing and deep learning.

\end{IEEEbiographynophoto}


\vfill


\end{document}